\RequirePackage{fix-cm}
\documentclass{article} 
\usepackage{iclr2027_conference,times}
\iclrfinalcopy

\usepackage{amsmath,amsfonts,bm}

\def\eqref#1{equation~\ref{#1}}

\def\1{\bm{1}}

\DeclareMathAlphabet{\mathsfit}{\encodingdefault}{\sfdefault}{m}{sl}
\SetMathAlphabet{\mathsfit}{bold}{\encodingdefault}{\sfdefault}{bx}{n}

\usepackage{soul,enumitem}
\usepackage{url}
\usepackage{amsmath,amssymb}
\usepackage{booktabs}
\usepackage{tabularx}
\usepackage{array}
\usepackage{multirow}
\usepackage{graphicx}
\usepackage{float}
\usepackage{wrapfig}
\usepackage{placeins}
\usepackage{microtype}
\usepackage{natbib}
\usepackage{xcolor}
\usepackage{listings}
\usepackage[most]{tcolorbox}
\usepackage{algorithm}
\usepackage{algpseudocode}
\usepackage{hyperref,cleveref}
\hypersetup{hidelinks}
\newcommand{\method}{\textsc{MeshHeal}}

\newcommand{\skillpair}[2]{\mbox{\texttt{#1} $+$ \texttt{#2}}}

\newcommand{\meshstate}[1]{\ifmmode\text{\bfseries #1}\else\textbf{#1}\fi}
\newcommand{\meshaction}[1]{\ifmmode\text{\bfseries #1}\else\textbf{#1}\fi}
\definecolor{meshblue}{HTML}{315F7D}
\definecolor{meshblueback}{HTML}{F4F8FB}
\definecolor{meshgreen}{HTML}{426B59}
\definecolor{meshgreenback}{HTML}{F4F8F5}
\definecolor{meshgray}{HTML}{5E666C}
\definecolor{meshgrayback}{HTML}{F7F7F7}

\lstdefinestyle{meshcode}{
  basicstyle=\ttfamily\fontsize{7.4}{9.0}\selectfont,
  breaklines=true,
  breakatwhitespace=false,
  columns=fullflexible,
  keepspaces=true,
  showstringspaces=false,
  morekeywords={normal,watched,isolated,Normal,Watched,Isolated,Handoff,Contribute,Finalize,Accept,Takeover},
  keywordstyle=\bfseries,
  tabsize=2
}
\newtcblisting{promptcard}[1]{
  enhanced jigsaw,
  breakable,
  listing only,
  listing engine=listings,
  listing options={style=meshcode},
  colback=meshblueback,
  colframe=meshblue,
  coltitle=white,
  fonttitle=\bfseries\small,
  title={#1},
  title after break={#1\ (continued)},
  boxrule=0.55pt,
  arc=1.2mm,
  left=1.4mm,
  right=1.4mm,
  top=1.2mm,
  bottom=1.2mm,
  before skip=5pt,
  after skip=5pt
}

\newtcblisting{tracecard}[1]{
  enhanced jigsaw,
  breakable,
  listing only,
  listing engine=listings,
  listing options={style=meshcode},
  colback=meshgrayback,
  colframe=meshgray,
  coltitle=white,
  fonttitle=\bfseries\small,
  title={#1},
  title after break={#1\ (continued)},
  boxrule=0.5pt,
  arc=1.2mm,
  left=1.4mm,
  right=1.4mm,
  top=1.2mm,
  bottom=1.2mm,
  before skip=4pt,
  after skip=4pt
}

\newtcolorbox{casecard}[1]{
  enhanced jigsaw,
  breakable,
  colback=meshgreenback,
  colframe=meshgreen,
  coltitle=white,
  fonttitle=\bfseries\small,
  title={#1},
  title after break={#1\ (continued)},
  boxrule=0.55pt,
  arc=1.2mm,
  left=1.8mm,
  right=1.8mm,
  top=1.4mm,
  bottom=1.4mm,
  before skip=5pt,
  after skip=5pt
}

\newtcolorbox{takeawaycard}[1]{
  enhanced jigsaw,
  breakable,
  colback=meshblueback,
  colframe=meshblue!75!black,
  coltitle=white,
  fonttitle=\bfseries\small,
  title={#1},
  title after break={#1\ (continued)},
  boxrule=0.45pt,
  arc=1.2mm,
  left=1.8mm,
  right=1.8mm,
  top=1.2mm,
  bottom=1.2mm,
  before skip=4pt,
  after skip=6pt
}

\crefname{figure}{Fig.}{Figs.}
\Crefname{figure}{Fig.}{Figs.}

\crefname{table}{Tab.}{Tabs.}
\Crefname{table}{Tab.}{Tabs.}

\crefname{section}{Sec.}{Secs.}
\Crefname{section}{Sec.}{Secs.}

\crefname{equation}{Eq.}{Eqs.}
\Crefname{equation}{Eq.}{Eqs.}

\crefname{algorithm}{Alg.}{Algs.}
\Crefname{algorithm}{Alg.}{Algs.}

\crefname{appendix}{App.}{App.}
\Crefname{appendix}{App.}{App.}

\title{
MeshHeal: Self-Healing Decentralized LLM Agent Networks under Gray Failures}

\author{
Keru Chen \\
School of ECEE \\
Arizona State University \\
\texttt{kchen234@asu.edu}
\And
Sen Lin \\
Computer Science Department \\
University of Houston \\
\texttt{slin50@central.uh.edu}
\And
Yingbin Liang \\
Department of ECE \\
The Ohio State University \\
\texttt{liang.889@osu.edu}
\AND
Nathaniel D. Bastian \\
Whiting School of Engineering \\
Johns Hopkins University \\
\texttt{ndbastian@jhu.edu}
\And
Shaofeng Zou\thanks{Corresponding author.} \\
School of ECEE \\
Arizona State University \\
\texttt{zou@asu.edu}
}

\begin{document}
\raggedbottom
\maketitle

\begin{abstract}
Decentralized LLM-based multi-agent systems coordinate through local interactions, but an agent can remain responsive while its task-solving quality persistently degrades. Such gray failures require protecting current tasks before sufficient evidence exists to alter future routing, while still allowing recovered agents to rejoin. We introduce \textsc{MeshHeal}, a fully decentralized self-healing framework that couples ability-matched peer review across two timescales. At the fast timescale, an adaptive hierarchy escalates uncertain or low-scoring outputs from repeated single-reviewer evaluation to committee deliberation and, when needed, correction before use. At the slow timescale, a task- and ability-conditioned peer-relative detector aggregates scores to distinguish persistent degradation from ordinary output variation, trigger mandatory committee review, and eventually exclude degraded agents from ordinary routing; recovery probes provide fresh evidence for reintegration. To faithfully evaluate routing, we introduce Model-Backed MAS Evaluation, which ties ability assignments to execution models, since prompt-based ability assignments alone can leave routing errors hidden. Across BBH, MATH, and MMLU-Pro, MeshHeal achieves 0.839 degraded-phase accuracy using 51k total model tokens per task, versus the strongest baseline Symphony’s 0.807 accuracy using 115k per task. Under staggered degradation and recovery, MeshHeal isolates degraded agents, keeps them excluded from ordinary task execution until recovery, and returns them to normal routing.

\end{abstract}

\section{Introduction}
\label{sec:introduction}

Recent advances in foundation models have enabled increasingly capable autonomous AI agents that can reason, plan, follow instructions, and use external tools \citep{bommasani2021opportunities}.
LLM-based multi-agent systems (MASs) decompose complex tasks among agents with complementary roles or abilities, combining their contributions into a final answer, often outperforming a single, larger model. However, most existing MASs coordinate in a centralized manner \citep{wu2024autogen,hong2024metagpt,qian2024chatdev,wang2025moa}, which creates a communication and scaling bottleneck and also a catastrophic single point of failure. 

Decentralized coordination instead allows agents to execute and route work through local interactions without relying on a central orchestrator. Accordingly, a growing body of recent work has investigated decentralized MAS architectures that distribute coordination and routing decisions across the agent network \citep{zhang2024coela,yang2025damcs,yang2025agentnet}. These studies, however, largely assume that participating agents remain reliable throughout execution. In practice, individual agents may fail or degrade because of problems in their model backends, local controllers, memories, or external tools, potentially disrupting task execution across the network \citep{huang2025resilience, zhang2025agent}.

\begin{figure}[!t]
\centering
\small
\includegraphics[width=\linewidth]{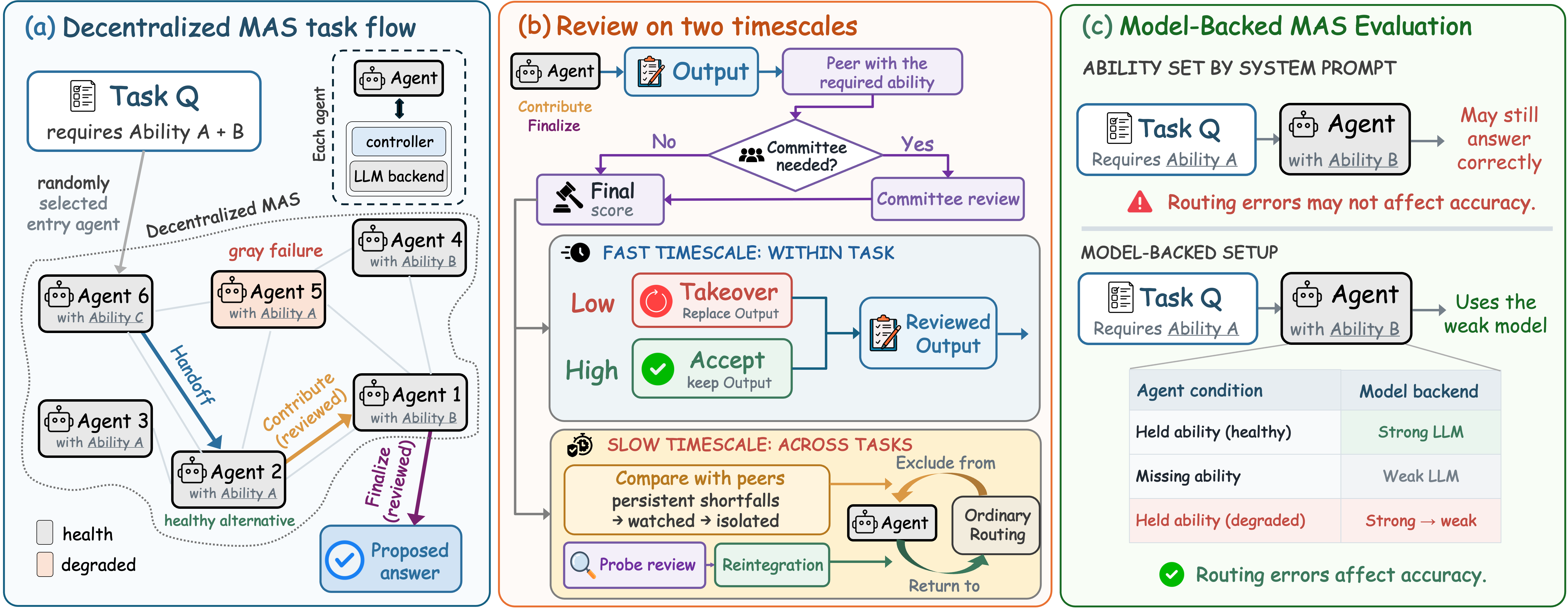}
\caption{Overview of \textbf{\method}. (a) Agents route and complete complementary work. (b) Review by agents with the relevant ability either keeps or replaces the current output, while scores accumulated across tasks update routing eligibility and probes allow recovered agents to return. (c) \textbf{Model-Backed MAS Evaluation} ties execution models to declared abilities and the agent's healthy or degraded condition, making routing errors affect accuracy.}
\label{fig:overview}
\end{figure}

Motivated by this gap, we study decentralized LLM-based MASs operating in the presence of agent failures. We focus on \emph{gray failures} \citep{huang2017gray,huang2018panorama}, and leave other types of failures as future work. 
A gray failure occurs when an agent remains available and protocol-compliant, i.e., it can communicate with its neighbors and return syntactically valid responses, but its effective ability to solve some tasks is degraded relative to its nominal ability. Such a failure is ``gray'' because it is not directly exposed through an explicit error or loss of connectivity: the affected agent may appear healthy to some neighbors, while its degradation can only be inferred from the semantic quality of its outputs or their downstream consequences.
Gray failures are particularly relevant to decentralized LLM-based multi-agent systems because an agent's availability does not necessarily imply its semantic reliability. 
We do not regard every incorrect LLM response as a gray failure, since even a
healthy agent has a nonzero probability of error. Instead, a gray failure
refers to a \textit{\textbf{sustained}} degradation relative to the agent's nominal task solving capability. Such degradation may arise from transient inference-serving issues, context corruption or overflow, stale memory, failures of external tools, or changes in the model backend. These failures are difficult to detect using conventional heartbeat- or connectivity-based mechanisms, since the affected agent continues to participate normally in the communication protocol.

We refer to the system-level capability to respond to such degradation as \emph{self-healing}: detecting agent degradation, limiting its effect on current task execution, adapting future routing, and reintegrating an agent after recovery \citep{psaier2011}.
Existing LLM inference and agent-coordination methods address complementary parts of this objective. Methods based on repeated inference, deliberation, or robust aggregation improve current outputs \citep{wang2023selfconsistency,du2024debate,li2024moreagents,wang2025moa,jo2025decentllms,zheng2025cpwbft,lee2026sac}, however, because they operate within individual tasks, these methods are not designed to retain producer-specific evidence over time and therefore do not support persistent-degradation detection, subsequent rerouting, or recovery-aware reintegration.
Moreover, history-based methods adapt routing across tasks \citep{li2026raps,yang2025agentnet}; however, routing adaptation alone acts only after history has accumulated, so it cannot protect the current task, and once an agent is bypassed, the system receives no fresh evidence of its recovery for reintegration. Therefore, to address gray failures, the system must protect the current task before degradation can be reliably detected, accumulate evidence to adapt future routing, and reintegrate an agent once its performance recovers.

We introduce \textbf{\method} (see \Cref{fig:overview}), a fully decentralized self-healing framework that couples peer review across two timescales. At the fast timescale, ability-matched neighboring agents review every intermediate contribution. An adaptive review hierarchy begins with repeated single-reviewer evaluation and escalates uncertain or low-scoring outputs to committee deliberation and, when necessary, multi-agent correction. At the slow timescale, a task- and ability-conditioned peer-relative detector accumulates the same review evidence to distinguish persistent degradation from ordinary output variation and impose progressively stronger routing restrictions, culminating in isolation from ordinary task execution. Recovery probes provide fresh reviewed evidence, allowing agents whose performance has recovered to return gradually to normal routing. 

In our experiments on BBH, MATH, and MMLU-Pro, \method{} achieves $83.9\%$ mean accuracy during degradation over five seeds, compared with $80.7\%$ for Symphony, the strongest baseline. It uses 51k tokens (both input and output tokens) per task, compared with 115k for Symphony and 178k for SAC, reducing token use by $56\%$ and $71\%$, respectively. In sparse networks of up to 576 agents, \method{} isolates all degraded agents and achieves comparable accuracy with a complete graph.

A commonly used approach to enforce ability constraints on agents when evaluating ability-aware routing is to use system prompts \citep{li2023camel,qian2024chatdev,chen2024agentverse,hong2024metagpt,chen2026hipo}. However, we identify an important confound: system prompts do not reliably enforce ability constraints, and the agent may still be able to solve tasks outside its abilities, allowing incorrect routes to produce correct final answers and thereby hiding routing errors. In a controlled BBH \citep{suzgun2023bbh} study across five models, we find that prompt-based ability assignments do not reliably separate performance on assigned and unassigned abilities.
We introduce Model-Backed MAS Evaluation, which uses backend selection to realize ability assignments and inject controlled gray failures. For a requested ability, a healthy agent uses a strong model if it holds that ability and a weak model otherwise.  This construction provides known failure ground truth while preserving the defining property that a gray-failed agent remains responsive.

\section{Related Work}
\label{sec:related_work}

\textbf{LLM multi-agent coordination and robustness.}
LLM agents combine reasoning and tool use to solve complicated tasks \citep{yao2023react}. MASs extend these capabilities through task decomposition and specialization \citep{wu2024autogen,hong2024metagpt,qian2024chatdev}, while decentralized systems coordinate and adapt through local interactions \citep{yang2025agentnet,li2026raps}. 

Various approaches have been proposed to improve output quality of LLM and MAS, e.g., repeated inference, deliberation, peer review, and robust aggregation \citep{wang2023selfconsistency,du2024debate,li2024moreagents,wang2025moa,jo2025decentllms,zheng2025cpwbft,lee2026sac}. These approaches may also be applied here to handle gray failure. However, they operate within each task and do not identify whether a persistent failure has occurred to an agent. Therefore, a degraded agent may consistently participate in task solving. 

Adaptive routing strategies were also developed to handle failures in MASs. AgentNet updates routing based on interaction history \citep{yang2025agentnet}, Symphony-Coord combines adaptive agent selection with voting \citep{guan2026symphony}, and RAPS uses reputation to identify and isolate unreliable peers \citep{li2026raps}. However, routing adaptation alone cannot protect the current task, and once an agent is bypassed, the system receives no fresh evidence of its recovery for reintegration.

\textbf{Distributed recovery and gray failures.}
Our recovery goal builds on self-stabilizing and self-healing systems, where systems adapt their behavior in response to failures \citep{dijkstra1974,arora1993,psaier2011}. Gray-failure research shows that hidden degradation may need to be inferred from peer observations rather than availability \citep{huang2017gray,huang2018panorama}. We study this problem for LLM agents using reviewed output quality as the failure signal.

\section{Problem Formulation}
\label{sec:problem}
 
We consider a decentralized MAS with heterogeneous agent abilities and potential gray failures.
\subsection{Decentralized MAS Architecture}

We model a decentralized LLM-based MAS as an undirected graph $G=(\mathcal V,\mathcal E)$, where $\mathcal V$ denotes the set of agents, and $\mathcal E\subseteq \mathcal V\times \mathcal V$ represents the communication graph between the agents:  an undirected edge $e_{i,j}\in\mathcal E$ allows agent $i$ to send information to agent $j$ and vice versa. Each agent consists of an LLM and a local controller. The LLM is used to generate the agent's outputs, and is called the agent's \emph{model backend}; and the local controller coordinates the agent's interactions with its neighbors.

We consider heterogeneous agents where agents differ in their abilities. Denote by $\mathcal{K}_i$ the set of abilities of an agent $i$, and $\mathcal{K}_i$ varies across agents. Solving different tasks may require a different set of abilities. Therefore, it is of key importance to assign the task to the agent with the matching ability. We study the decentralized setting, where no agent has a global view of other agents, and each agent can only communicate with its neighbors on the communication graph $\mathcal E$.

The multi-agent system will be given a sequence of tasks denoted by $\mathcal T=\{\tau_1,\tau_2,\ldots, \tau_N\}$.
The $t$-th task is $\tau_t=(Q_t,T_t)$, where $t\in[N]$ is the task index, $Q_t$ is the input question, $T_t$ is its benchmark-provided category label (e.g., \texttt{geometry} in MATH \citep{hendrycks2021math}), and we use $R_t$ to represent the set of abilities required to solve $Q_t$. Here $[N]\triangleq\{1,\ldots,N\}$. We note that $Q_t$ and $T_t$ are directly given to the agent, but $R_t$ needs to be inferred by the agent.

\textbf{Objective.} Let $\Pi_{\mathrm{local}}$ be the set of coordination policies whose decisions use only the current task state, local interaction history, and messages from direct neighbors. For a policy $\pi$, let $\widehat y_t^{\pi}$ be the final answer of the MAS and $\operatorname{Eval}_t(\widehat y_t^{\pi})\in\{0,1\}$ indicate correctness under the benchmark evaluator. We seek to optimize the average accuracy over the $N$ sequential tasks:
\begin{equation}
\max_{\pi\in\Pi_{\mathrm{local}}}
\frac{1}{N}\sum_{t=1}^{N}\mathbb E\!\left[\operatorname{Eval}_t(\widehat y_t^{\pi})\right],
\label{eq:objective}
\end{equation}
where the expectation is over the execution randomness. 

\subsection{Gray Failure and Self-Healing}
We study a practical setting in which agents may experience \emph{gray failures} \citep{huang2017gray,huang2018panorama}. 
A gray failure occurs when an agent remains available and protocol-compliant, i.e., it can communicate with its neighbors and return syntactically valid responses, but its effective ability to solve some tasks is degraded relative to its nominal ability. 

Each agent may have two possible health conditions: \textit{healthy}, \textit{degraded}. The health condition of each agent $i$ may change over time. We assume that each agent's health condition remains unchanged within a task but may change across tasks. 
An agent is called \emph{healthy} when it operates at its normal level of reliability, and \emph{degraded} when it suffers from gray failure. The true health condition is latent,  and the agent must infer it from the agent's outputs. 
We further define \textit{recovery} as restoration of an agent from \textit{degraded} to \textit{healthy}.

In this paper, our focus is on solving the problem in \Cref{eq:objective} when agents experience gray failures and may also recover from them. We aim to design local coordination policies so that, even with degraded agents, each task can still be routed to healthy agents with matching abilities and solved with high accuracy; and at the same time, when an agent recovers from \textit{degraded} to \textit{healthy}, future tasks that match this agent's abilities will still be routed to it. Such desired property is referred to as \textbf{\textit{self-healing}}.
To evaluate this ability, beyond task accuracy and inference cost, we introduce three additional criteria: detection delay, closure, and reintegration. These criteria are adapted from the distributed fault-tolerance literature \citep{dijkstra1974,arora1993,ghosh2007,psaier2011}.
More specifically, suppose agent $i$ becomes degraded at task $b_i\in[N]$ and recovers at task $r_i\in[N+1]$. If recovery is not observed in an $N$-task stream, we set $r_i=N+1$. Let $\mathcal I_t$ denote the set of agents that are excluded from ordinary task execution at task $t$. Then, we define:

\begin{itemize}[leftmargin=*, nosep]
\item \textit{Detection delay:} For a degraded agent $i$, the system should isolate it as early as possible. Let $c_i=\min\{t\ge b_i:i\in\mathcal I_t\}$ be the index of the first task at which the degraded agent $i$ is isolated. 
The detection delay is then defined as $\delta_i^{\mathrm{delay}}=c_i-b_i$.
\item \textit{Closure:} A system satisfies the \textit{closure} property if a degraded agent remains isolated until recovery after being detected. Formally, for any degraded agent $i$,  $i\in \mathcal I_t$,  $\forall c_i\le t<r_i$.
\item \textit{Reintegration:} it refers to the case that after recovery, agent $i$ returns to normal routing. Let $u_i\ge r_i$ be the index of the first task at which agent $i$ returns to normal routing state through the end of the observed stream. Then the reintegration delay is defined as $\delta_i^{\mathrm{rein}}=u_i-r_i$.
\end{itemize}

\section{\method{}: Review at Two Timescales}
\label{sec:method}

In \Cref{sec:base}, we will present our base workflow design in a decentralized MAS \textit{without gray failures}. Then, in \Cref{sec:4.2}, we build upon the base design and introduce our \method{} with novel two-timescale mechanism for self-healing. The full algorithm is provided in \Cref{alg:meshheal} in App.~\ref{app:algorithm}.

\subsection{Base Task Execution and Routing (No Failures)}\label{sec:base}

\textbf{Task execution.}
Agents collaborate  to sequentially solve and route a task through the network. Each task $\tau_t=(Q_t,T_t)$ is initially assigned to an entry agent sampled uniformly at random from the network. The query $Q_t$ and task type $T_t$ are given, while $R_t$ denotes the set of abilities required to solve the task and is inferred by the entry agent. 

The task is solved sequentially as it moves through the network, with each agent addressing the part aligned with its abilities and passing the rest to the next agent until it is fully solved.
The task state is initialized as $X_t=(Q_t,T_t,R_t,C_t,Y_t,P_t)$, where $C_t$ records the abilities already covered so far, $Y_t$ stores the working context that will be passed to later agents, including accepted intermediate outputs, and $P_t$ is a list that records the route taken so far. Initially, $C_t=Y_t=\emptyset$ and $P_t$ is also empty. 

When the task is routed to agent $i$, let $M_t=R_t\setminus C_t$ be the set of abilities that were not supplemented by previous agents in $P_t$ and need to be further provided by subsequent agents, and $H_{i,t}=\mathcal K_i\cap M_t$ be the subset of matching abilities that can be provided by agent $i$. We take a rule-based approach at the controller, and these sets govern the controller's decision for the current task:
\begin{itemize}[leftmargin=*, nosep]
    \item If $H_{i,t}=\emptyset$, then agent $i$ has no matching abilities to contribute to the remaining of task $t$. Then, agent $i$'s controller chooses to \meshaction{Handoff}: it generates no task content and forwards the task state unchanged to the next agent (discussed later in Routing). Thus, $C_t$, $Y_t$ and $P_t$ remain unchanged.
    
    \item If $\emptyset\neq H_{i,t}\subset M_t$, then agent $i$ contribute matching abilities to the task that prior agents in $P_t$ cannot; however, additional abilities beyond those in $H_{i,t}$ are still needed. Then, agent $i$'s controller chooses to 
    \meshaction{Contribute}. The agent's LLM solves the matching part in the remaining of the task using abilities in $H_{i,t}$. The output is appended to $Y_t$, $C_t=C_t\cup H_{i,t}$, and agent $i$ is appended to the list $P_t$. The controller recomputes $M_t \leftarrow R_t \setminus C_t$ using the updated $C_t$ and forwards the updated task state to the next agent (discussed later in Routing).
    \item If $H_{i,t}=M_t$, then agent $i$ has all the abilities needed to finish the remaining of the task. Then, agent $i$'s controller chooses to 
    \meshaction{Finalize}. The agent solves the remaining task, combines the result with earlier outputs in $Y_t$, and proposes the final answer. At this point, all abilities in $R_t$ have been covered by agents along the route in $Y_t$, and no further routing is required.
\end{itemize}

\textbf{Routing.} Each agent $i$ maintains an estimated the shortest hop distance $D_i(s)$ to the nearest agent that holds ability $s$. If agent $i$ itself holds ability $s$, then $D_i(s)=0$; otherwise, it updates in a distributed manner:
$
D_i(s)=1+\min_{j\in\mathcal N_i} D_j(s),
$
where $\mathcal N_i$ is the set of direct neighbors of agent $i$ \citep{bellman1958routing}. Under standard asynchronous Bellman-Ford assumptions with fixed topology and reliable neighbor updates, these iterations converge in finite time \citep{bertsekas1982distributed}. The path that achieves $D_i(s)$ will be used to select the next agent.
When \meshaction{Handoff} or \meshaction{Contribute} are chosen by the controller, the task has at least one ability not provided along the route in $P_t$. The controller samples one ability $s$ uniformly at random from $M_t$. It then routes the task state along the path that achieves $D_i(s)$, i.e., to the nearest agent with ability $s$. 

\subsection{Task Execution and Routing under Gray Failures}\label{sec:4.2}

Having defined the base task flow without failures, we now present our novel design to address gray failures. We define a state
$h_i\in\{\meshstate{normal},\meshstate{watched},\meshstate{isolated}\}$ for each agent $i$, which determines how the agent participates in execution and routing. A \meshstate{normal} agent follows the base workflow without restriction. 
A \textbf{watched} agent is routed and executes tasks as a normal agent, but every output it produces receives mandatory committee review. An \meshstate{isolated} agent has its set of abilities $\mathcal K_i$ set to be $\emptyset$ temporarily, and therefore it will be excluded from task execution, but its controller remains active for message relay.
When an agent becomes \meshstate{isolated}, agents will need to recompute the shortest hop distance as described in \Cref{sec:base}. We then apply the base task flow in \Cref{sec:base}.

Our \method{} framework presents a novel two-timescale scheme with fast review and correction and slow detection, rerouting, and recovery (illustrated in \Cref{fig:overview}). At the \textit{fast timescale}, an review protocol escalates low-scoring, high-disagreement, or watched-agent outputs from repeated single-reviewer evaluation to deliberative committee review and multi-agent correction. At the \textit{slow timescale}, a task- and ability-conditioned peer-relative detector aggregates persistent shortfalls and normalizes them against healthy-neighbor variation, enabling same thresholds to be used across model backends; guarded recovery probes close the loop through reintegration.

\subsubsection{Fast Timescale: Review and Correction}\label{sec:fast}
\label{sec:self_healing}

At the fast timescale, every output produced by \meshaction{Contribute} or \meshaction{Finalize} is reviewed before use. We denote the output by $y_{i,t}$. An eligible reviewer must be a \meshstate{normal} direct neighbor that holds every ability in $H_{i,t}$. The reviewer scores $y_{i,t}$ on a fixed $[0,1]$ scale (judge prompts in App.~\ref{app:review_prompt}).
To reduce review noise, the same reviewer evaluates $y_{i,t}$ in three independent calls. Denote the median by $q_0$, and we use it as the initial score. We use the median rather than the mean to reduce sensitivity to reviewer variability and outlier scores, consistent with prior work on robust aggregation of LLM judgments \citep{acharya2026ropoll,gallaba2025towards}.
We also use the standard deviation $s_u$ to measure disagreement. Committee review is triggered when $q_0\le\tau_{\mathrm{to}}$, $s_u\ge\sigma_{\mathrm{com}}$, $h_i=\meshstate{watched}$, or the output is a recovery probe, where $\tau_{\mathrm{to}}=0.40$ is the acceptance threshold and $\sigma_{\mathrm{com}}=0.55$ is the disagreement threshold. 
Committee review adds multiple perspectives and deliberation, improving review reliability at the cost of additional inference and communication; we therefore invoke it only for low-scoring, high-disagreement, or watched-agent outputs.

The committee contains two eligible reviewers. They first evaluate $y_{i,t}$ independently, exchange their explanations, and then reassess the same output. Let $v_A$ and $v_B$ be their second-round scores; the final review score is $q_{i,t}=(v_A+v_B)/2$. If committee review is not triggered, $q_{i,t}=q_0$. When only one eligible reviewer is available, two reviews are obtained through separate inference calls with isolated context histories.
If $q_{i,t}>\tau_{\mathrm{to}}$, $y_{i,t}$ is accepted. Otherwise, \meshaction{Takeover} is triggered: the two reviewers generate correction candidates, which a committee reviewer in the normal state synthesizes task context following Mixture-of-Agents style aggregation \citep{wang2025moa}. The resulting corrected output $\widehat y_{i,t}$ replaces $y_{i,t}$ before entering the task state. The score $q_{i,t}$ remains attributed to agent $i$ and is retained as evidence for the slow timescale degradation detection.

\subsubsection{Slow Timescale: Detection, Rerouting and Recovery}\label{sec:slow}

A single low score in the fast timescale does not necessarily mean that the agent is persistently degraded. We therefore accumulate the review scores $q_{i,t}$ across tasks and use \emph{relative peer comparison} to update the estimate of $h_i$. Our goal is to detect agent gray failures across different task types and model backends with the same detector, rather than tuning a separate threshold for each setting. 

\textbf{Relative shortfall.}
For each ability $s\in H_{i,t}$, the controller considers direct \meshstate{normal} neighbors that have recently reviewed outputs within the fixed history window from earlier tasks with the same task type $T_t$ and ability $s$. Let $\mathcal P_{i,T_t,s}(t)$ denote this set of neighbors, excluding $i$, and let $\mu_{j,T_t,s}(t)$ denote neighbor $j$'s recent mean review score before task $t$. The controller first uses these neighbors to estimate the normal review score for the same type of work $(\bar q_{-i,T_t,s}(t))$. For each $s\in H_{i,t}$, the controller measures how far agent $i$'s current score falls below the corresponding reference:
$$
\bar q_{-i,T_t,s}(t)
=
\frac{1}{|\mathcal P_{i,T_t,s}(t)|}
\sum_{j\in\mathcal P_{i,T_t,s}(t)}
\mu_{j,T_t,s}(t),
\qquad
x_{i,t,s}
=
\bar q_{-i,T_t,s}(t)-q_{i,t}.
$$
A positive $x_{i,t,s}$ means that agent $i$ scored below \meshstate{normal} neighbors on the same task type and ability $s$. This comparison removes differences in absolute review scores across task types by comparing an agent only with normal neighbors who work on the same task type and have the same ability.

\textbf{Persistent degradation.}
A single positive shortfall may result from LLM randomness. We therefore maintain a separate detector for each $(T,s)$ and average its most recent 8 shortfalls, yielding $\bar x_{i,T,s}$. For a healthy agent, it should stay near zero; if the agent remains degraded, it should stay positive.

However, a fixed threshold on \(\bar x_{i,T,s}\) may not transfer across model backends because the dispersion of healthy agents’ mean shortfalls can differ substantially across backends. We therefore also measure how much the mean shortfalls of \meshstate{normal} neighbors vary. We pool the mean shortfalls $\bar x_{j,T,s'}$ of all \meshstate{normal} neighbors $j$ over all abilities $s'$ under task type $T$, and let $m_{i,T}$ and $p^{25}_{i,T}$ be their median and 25th percentile. We measure the normal variation by the lower spread $m_{i,T}-p^{25}_{i,T}$, because degraded agents that have not yet been detected can inflate the upper tail. We normalize agent $i$'s mean shortfall by this variation:
$$
z_{i,T,s}=\frac{\bar x_{i,T,s}-m_{i,T}}{\max\!\left(m_{i,T}-p_{i,T}^{25},\,0.10\right)
}.
$$ 
Thus, $z_{i,T,s}$ measures how far agent $i$'s mean shortfall is above normal level, relative to how much healthy agents vary. This second comparison accounts for differences in how much healthy agents' mean shortfalls vary across model backends, so same detection thresholds can be used across models.

\textbf{State transitions, rerouting, and recovery.}
Each update uses the current $z_{i,T,s}$; if fewer than three matched observations are available $h_i$ remains unchanged. Otherwise, $z_{i,T,s}\ge\alpha$ moves \meshstate{normal} to \meshstate{watched}, $z_{i,T,s}<\alpha$ returns \meshstate{watched} to \meshstate{normal}, and two consecutive updates with $z_{i,T,s}\ge\beta$ move a \meshstate{normal} or \meshstate{watched} agent to \meshstate{isolated}; the isolation count resets when $z_{i,T,s}<\beta$. Here $\alpha <\beta$ are thresholds, and in our experiments we choose $\alpha=1.2$ and $\beta = 2$. The thresholds were selected empirically to balance detection delay and false isolation of healthy agents.

We further design \emph{recovery probes} that use the same review and detection mechanism. After every 10 tasks, an isolated agent becomes eligible for a recovery probe. When a neighbor next receives work matching its abilities, the work is routed to the isolated agent and its output receives mandatory committee review. The resulting score updates the same detector. The agent returns to \meshstate{watched} after one update with $z_{i,T,s}<\beta$, even if
$z_{i,T,s}<\alpha$, and to \meshstate{normal} after a later update with $z_{i,T,s}<\alpha$.

\section{Model-Backed MAS Evaluation}                                                        
\label{sec:dataset}

\begin{wraptable}{r}{0.55\linewidth}
\vspace{-7pt}
\centering
\caption{BBH ability-assignment test. Arrows show no competence statement $\rightarrow$ unassigned abilities declared unreliable; $\Delta_{\mathrm{gap}}$ is the resulting increase in the accuracy gap between assigned and unassigned abilities.}
\label{tab:prompt_only_main}
\vspace{-2pt}
\fontsize{7.4}{8.6}\selectfont
\setlength{\tabcolsep}{2.5pt}
\begin{tabular}{@{}lccc@{}}
\toprule
Model & \shortstack{Unassigned\\ability acc.} & \shortstack{Assigned\\ability acc.} & $\Delta_{\mathrm{gap}}$ \\
\midrule
DeepSeek-Chat & $.812\!\rightarrow\!.775$ & $.800\!\rightarrow\!.792$ & .029 \\
Llama-3.1-70B & $.781\!\rightarrow\!.750$ & $.783\!\rightarrow\!.767$ & .015 \\
GPT-4o-mini & $.744\!\rightarrow\!.675$ & $.708\!\rightarrow\!.667$ & .027 \\
GPT-OSS-120B & $.869\!\rightarrow\!.831$ & $.825\!\rightarrow\!.833$ & .046 \\
Qwen-2.5-7B & $.494\!\rightarrow\!.219$ & $.558\!\rightarrow\!.342$ & .058 \\
\bottomrule
\end{tabular}
\vspace{-3pt}
\end{wraptable}
\textbf{System prompt fails to constrain the agent's ability.}
Many LLM MAS evaluations assign abilities through system prompts while using the same underlying model for all agents \citep{li2023camel,qian2024chatdev,chen2024agentverse,hong2024metagpt}. 
To test whether such prompts constrain agents, we compare two prompting conditions across five models on BBH benchmark \citep{suzgun2023bbh}: one makes no statement about unassigned abilities, while the other explicitly describes them as unreliable. If these prompts truly limited agents to their assigned abilities, accuracy on unassigned abilities should fall while assigned-ability accuracy remains stable.
The accuracy gap between assigned and unassigned abilities increases by only $0.015$--$0.058$ (see \Cref{tab:prompt_only_main}). Accuracy on unassigned abilities remains substantial, while assigned-ability accuracy also falls for four models. Prompt instructions change agent behavior, but do not reliably prevent agents from solving tasks outside their assigned abilities. App.~\ref{app:competence} gives the full protocol.

\textbf{Controlled ability realization and gray-failure injection.}
We use model backends to realize ability assignments and inject gray failures, shown in \Cref{fig:overview}(c). A healthy agent uses a \textbf{strong model} for assigned abilities and a \textbf{weak model} otherwise. Degradation is simulated by replacing its strong backend with the weak one while leaving communication and execution interfaces unchanged. This preserves responsiveness while reducing task competence, providing known failure ground truth and making routing errors visible in end-to-end accuracy.

\textbf{Tasks and degraded agents.}
We apply this protocol to seven task groups from each of BBH, MATH, and MMLU-Pro \citep{suzgun2023bbh,hendrycks2021math,wang2024mmlupro}. Each task group contains tasks of the same type and is associated with two complementary abilities required for solving those tasks. The two abilities are assigned to disjoint sets of agents, so no single agent holds both. Every task requires at least two agents. Full task-to-ability mappings appear in App.~\ref{app:implementation}.

\section{Experiments}
\label{sec:experiments}

We evaluate on benchmarks of BBH, MATH, and MMLU-Pro. The main system contains six agents, A1--A6, with three agents per ability and disjoint groups for the two abilities required by each task. A3 and A5 are selected for degradation, every LLM call by them, including review calls, switches to the weak model. Their controllers, abilities, communication links, and availability remain unchanged. Unless stated otherwise, healthy agents use \texttt{openai/gpt-oss-120b} for declared abilities; undeclared abilities and degraded agents use \texttt{meta-llama/llama-3.2-1b-instruct}.

We compare \method{} with AgentNet's decentralized routing \citep{yang2025agentnet}, RAPS's reputation-based routing \citep{li2026raps}, and a centralized AutoGen implementation with a mandatory manager \citep{wu2024autogen}. SAC uses three teams with two rounds of filtering and refinement followed by majority aggregation \citep{lee2026sac}, while Symphony uses parallel plans and voting \citep{guan2026symphony}. All methods use the same tasks, agents, ability assignments, models, degradation targets, and schedule. In AutoGen, A3 is the mandatory manager and appears on every task path.

Each setting uses five seeds, defining task samples and order shared across methods. We report means and sample standard deviations. Except in targeted ablations, review rubrics and detector parameters are fixed across datasets, model pairs, and network sizes (\Cref{tab:parameters}). We measure accuracy, token consumption (counting both input and output tokens) , detection delay $\delta_i^{\mathrm{delay}}$, closure, and reintegration delay $\delta_i^{\mathrm{rein}}$; Appendices~\ref{app:implementation}--\ref{app:cases} provide settings, prompts, traces, and full results.

\subsection{Persistent Degradation: Accuracy and Cost}

Each stream contains 252 healthy-phase tasks followed by 252 disjoint degraded-phase tasks with the same task-type distribution. During degraded-phase, A3 and A5 switch to the weak model. We report healthy- and degraded-phase accuracy and total model-token cost. After degradation, \method{} reaches $0.839\pm0.012$ accuracy at $51\pm5$k total model tokens per task, compared with Symphony's $0.807\pm0.014$ at $115\pm9$k. AgentNet and RAPS use fewer tokens but have substantially lower accuracy (\Cref{tab:main_and_scale}(a)). Full dataset-level and cost results are in App.~\ref{app:results}.

\begin{table*}[t]
\centering
\caption{
(a) Accuracy and total model-token cost under persistent degradation
(mean $\pm$ std.\ over three datasets, each with five seeds).
H/D denote healthy/degraded phases; tokens are thousands per degraded-phase task.
(b) Scaling in complete and sparse networks over five seeds.
$I/D$ denotes isolated/degraded agents; edge fraction is relative to the complete graph.
}
\small
\setlength{\tabcolsep}{1.7pt}
\vspace{2pt}
\begin{minipage}[t]{0.46\textwidth}
\centering

\textbf{(a) Persistent degradation}
\vspace{2pt}

\small
\renewcommand{\arraystretch}{0.95}
\begin{tabular}{@{}lccc@{}}
\toprule
Method & H acc. & D acc. & Tokens \\
\midrule
\textbf{\method}
& $\mathbf{.831\!\pm\!.013}$
& $\mathbf{.839\!\pm\!.012}$
& $51\!\pm\!5$ \\
Symphony
& $.827\!\pm\!.013$
& $.807\!\pm\!.014$
& $115\!\pm\!9$ \\
SAC
& $.822\!\pm\!.014$
& $.746\!\pm\!.021$
& $178\!\pm\!13$ \\
RAPS
& $.800\!\pm\!.019$
& $.653\!\pm\!.041$
& $48\!\pm\!3$ \\
AgentNet
& $.729\!\pm\!.023$
& $.557\!\pm\!.039$
& $43\!\pm\!2$ \\
AutoGen
& $.787\!\pm\!.014$
& $.424\!\pm\!.023$
& $15\!\pm\!0$ \\
\bottomrule
\end{tabular}
\end{minipage}
\hfill
\begin{minipage}[t]{0.50\textwidth}
\centering
\vspace{2pt}
\textbf{(b) Network scaling}
\vspace{6pt}

\setlength{\tabcolsep}{2.5pt}
\renewcommand{\arraystretch}{0.95}
\begin{tabular}{@{}rccccc@{}}
\toprule
& \multicolumn{2}{c}{Complete}
& \multicolumn{2}{c}{Sparse}
& Edge \\
\cmidrule(lr){2-3}\cmidrule(lr){4-5}
Agents & Acc. & $I/D$ & Acc. & $I/D$ & fraction \\
\midrule
96  & .827 & 32/32   & .818 & 32/32   & 5.00\% \\
288 & .813 & 96/96   & .803 & 96/96   & 1.71\% \\
576 & .808 & 192/192 & .814 & 192/192 & 0.88\% \\
\bottomrule
\end{tabular}
\end{minipage}


\label{tab:main_and_scale}
\end{table*}

\subsection{Degradation and Recovery: Detection, Closure, and Reintegration}

We test the three self-healing criteria on five consecutive 126-task MATH phases: healthy operation, A5 degraded, A3 and A5 degraded, A3 degraded after A5's backend restoration, and both backends restored. All methods complete the 630-task stream. Across the 504 tasks after the first degradation, including the fully restored phase, \method{} reaches $0.812$ accuracy at 49k total model tokens per task, compared with Symphony's $0.806$ at 105k (shown in \Cref{tab:staggered}). Symphony leads in the A5-only and fully recovered phases, while the methods tie when only A3 remains degraded.

\begin{table*}[!t]
\centering

\caption{
Self-healing under staggered degradation and recovery over five seeds.
(a) Mean accuracy and total model-token cost across phases.
(b) Detection (Det.) and reintegration (Rein.) delays are mean $\pm$ std.\
(min--max); closure reports successful seeds.
}
\small

\makebox[\textwidth][c]{%
\begin{minipage}[t]{0.59\textwidth}
\centering
\vspace{0pt}
\textbf{(a) Accuracy and token cost}\\[1pt]

\fontsize{8pt}{9.5pt}\selectfont
\setlength{\tabcolsep}{2.2pt}
\renewcommand{\arraystretch}{0.92}
\begin{tabular}{@{}lrrrrrr@{}}
\toprule
Phase & \method & Sym. & SAC & A.Net & RAPS & AutoGen \\
\midrule
Healthy              & .833 & .810 & .825 & .794 & .722 & .627 \\
A5 deg.               & .778 & .786 & .722 & .579 & .492 & .659 \\
A3+A5 deg.            & .770 & .722 & .643 & .452 & .405 & .183 \\
A5 restored; A3 deg.  & .841 & .841 & .706 & .532 & .524 & .135 \\
All restored          & .857 & .873 & .860 & .563 & .524 & .770 \\
\midrule
Pooled after deg. &
\textbf{.812} & .806 & .733 & .532 & .486 & .437 \\
Tokens/task after deg. &
49k & 105k & 124k & 39k & 38k & 15k \\
\bottomrule
\end{tabular}
\end{minipage}
\hspace{0.018\textwidth}
\begin{minipage}[t]{0.35\textwidth}
\centering
\vspace{5pt}
\textbf{(b) Self-healing metrics of \method{}}\\[1pt]
\vspace{5pt}
\fontsize{7pt}{11.5pt}\selectfont
\setlength{\tabcolsep}{2.0pt}
\renewcommand{\arraystretch}{1.3}
\begin{tabular}{@{}lccc@{}}
\toprule
Agent & Det.$\downarrow$ & Closure & Rein.$\downarrow$ \\
\midrule
A5 &
\shortstack{$8.0\pm5.2$\\[-1pt](2--16)} &
5/5 &
\shortstack{$46.0\pm11.7$\\[-1pt](30--63)} \\
A3 &
\shortstack{$12.4\pm6.2$\\[-1pt](6--22)} &
5/5 &
\shortstack{$45.0\pm3.5$\\[-1pt](42--51)} \\
\bottomrule
\end{tabular}
\end{minipage}%
}

\label{tab:staggered}
\end{table*}

Across five seeds, after agent $i$ becomes degraded at $b_i$, it is first excluded from ordinary task execution at $c_i$. The resulting detection delays $\delta_i^{\rm delay}=c_i-b_i$ are $8.0\pm5.2$ tasks for A5 and $12.4\pm6.2$ tasks for A3. Once isolated, both agents satisfy closure in all five seeds, remaining excluded throughout $c_i \le t < r_i$ except for committee-reviewed recovery probes. After backend restoration at $r_i$, the agents return to \meshstate{normal} routing at $u_i$, giving reintegration delays $\delta_i^{\rm rein}=u_i-r_i$ of $46.0\pm11.7$ tasks for A5 and $45.0\pm3.5$ tasks for A3 (\Cref{tab:staggered}). \Cref{fig:detect_recover} in App.~\ref{sec:model_and_recovery} shows one representative trajectory, with its state changes listed in \Cref{tab:state_trajectory}. None of the baseline methods fully isolates degraded agents; under our definitions, their detection delay is therefore infinite, closure is not established, and reintegration delay is undefined.

\subsection{Scaling with Local Information}

We replicate the six-agent profiles in complete and sparse graphs with 96, 288, and 576 agents, with one third degraded and all review, summary exchange, and routing updates remaining local. \method{} isolates every degraded agent in all settings, while sparse and complete accuracies differ by at most $0.010$. At 576 agents, the sparse graph uses only $0.88\%$ of complete-graph edges and reaches $0.814$ accuracy (\Cref{tab:main_and_scale}(b)), demonstrating \method{}'s effectiveness in sparse networks.

\subsection{Component and Detector Studies}

\textbf{Fast timescale.} We replace committee review with a single reviewer and separately remove Takeover on BBH while retaining detection and rerouting (\Cref{tab:mechanisms}(a); App.~\ref{app:ablations}, \Cref{tab:component_ablation}). Single reviewer uses the initial peer for judgment and correction; it has a slightly higher healthy-phase mean but $0.035$ lower degraded-phase accuracy and a mean phase change of $-0.044$. No Takeover leaves low-scoring outputs unchanged, reducing degraded-phase accuracy by $0.056$ and healthy-phase accuracy by $0.020$. This suggests that Takeover also corrects some errors from healthy agents. An audit (App.~\ref{app:review_audit}, \Cref{tab:agent_outcomes}(a,b)) shows that healthy agents receive much higher review scores than degraded agents ($0.89$ vs.\ $0.08$), and $98\%$ of degraded outputs are sent to committee review. Committee reassessment reduces false alarms on healthy agents. A separate Takeover audit (\Cref{tab:agent_outcomes}(c)) shows that Takeover corrects 15 of 32 wrong answers while turning only 1 of 32 correct answers into an incorrect one.

\begin{table}[!t]
\centering
\small
\caption{
(a) Accuracy over five seeds on BBH; H and D denote the healthy and degraded phases.
(b) Missed degraded agents out of two and falsely \meshstate{isolated} healthy agents out of four on BBH.
}
\label{tab:mechanisms}
\vspace{-2pt}

\makebox[\linewidth][c]{%
\begin{minipage}[t]{0.48\linewidth}
\centering
\vspace{0pt}
\textbf{(a) Fast-timescale components}\\[2pt]

\fontsize{7.4}{8.1}\selectfont
\setlength{\tabcolsep}{1.5pt}
\renewcommand{\arraystretch}{0.95}
\begin{tabular}{@{}lrrr@{}}
\toprule
Setting & H & D & $\Delta$ \\
\midrule
Full
& $.899\pm.027$
& $\mathbf{.900\pm.014}$
& $\mathbf{+.001\pm.016}$ \\
Single reviewer
& $\mathbf{.909\pm.022}$
& $.865\pm.029$
& $-.044\pm.024$ \\
No Takeover
& $.879\pm.021$
& $.844\pm.036$
& $-.035\pm.030$ \\
\bottomrule
\end{tabular}
\end{minipage}
\hfill
\begin{minipage}[t]{0.50\linewidth}
\centering
\vspace{-3pt}
\textbf{(b) Detector comparison}\\[2pt]

\fontsize{7.4}{8.1}\selectfont
\setlength{\tabcolsep}{2.0pt}
\renewcommand{\arraystretch}{0.95}
\begin{tabular}{@{}lcccc@{}}
\toprule
& \multicolumn{2}{c}{Pair A}
& \multicolumn{2}{c}{Pair B} \\
\cmidrule(lr){2-3}\cmidrule(lr){4-5}
Rule & Miss & False & Miss & False \\
\midrule
Absolute threshold
& 0/2 & 4/4 & 0/2 & 0/4 \\
CUSUM
& 0/2 & 2/4 & 1/2 & 0/4 \\
Relative peer comparison
& 0/2 & 0/4 & 0/2 & 0/4 \\
\bottomrule
\end{tabular}
\end{minipage}%
}
\end{table}

\textbf{Slow timescale.} We compare detectors under Pair A (GPT-OSS-120B/Gemma-3-4B) and Pair B (Llama-3.3-70B/Llama-3.2-1B). The absolute threshold falsely isolates all four healthy agents under Pair A; CUSUM \citep{page1954cusum} falsely isolates two under Pair A and misses one of two degraded agents under Pair B. Relative peer comparison detects both degraded agents without false isolations under either pair using fixed rubrics and parameters (\Cref{tab:mechanisms}(b); App.~\ref{app:ablations}, \Cref{tab:detector_transfer}).

\textbf{Model-pair sensitivity.}
To test robustness under a milder degradation and a weaker healthy model, we additionally evaluate Pair A and Pair B, respectively. \method{} detects both degraded agents without false isolation across both settings, while maintaining competitive degraded-phase accuracy at substantially lower cost: it outperforms Symphony by $0.040$ on Pair A and is within $0.004$ on Pair B, using only $52$--$56\%$ as many total model tokens. Full results are in App.~\ref{sec:model_and_recovery}.

\vspace{-3pt}
\section{Conclusion}
\label{sec:conclusion}

We introduced \method{}, a fully decentralized framework for self-healing gray failures in LLM agent networks with a novel two-timescale review design. At the fast timescale, review protocol evaluates every produced output and can replace low-quality outputs before they enter the task state; and at the slow timescale, a task- and ability-conditioned peer-relative detector aggregates persistent shortfalls to drive staged state transitions and routing updates, while recovery probes support evidence-based reintegration. We further showed that prompt-only ability assignments do not reliably enforce constraints on agents' ability and proposed Model-Backed MAS Evaluation, which uses controlled backend selection to realize complementary abilities and inject gray failures with known ground truth while keeping affected agents responsive.
With the default model pair, \method{} reaches $0.839$ degraded-phase accuracy at 51k tokens per task across three benchmarks and five seeds, versus the strongest baseline of Symphony  $0.807$ at 115k tokens per task. The experiment with degradation and recovery at different times meets the three self-healing criteria under the evaluated schedule. The scaling experiments isolate every degraded agent in the evaluated networks of up to 576 agents.

\clearpage
\bibliography{iclr2027_conference}
\bibliographystyle{iclr2027_conference}

\newpage
\appendix
\makeatletter
\@addtoreset{table}{section}
\@addtoreset{figure}{section}
\makeatother
\renewcommand{\thetable}{\Alph{section}\arabic{table}}
\renewcommand{\thefigure}{\Alph{section}\arabic{figure}}

\section{Ability Assignment Through System Prompts}
\label{app:competence}

We test whether an ability assignment in the system prompt can create a reliable performance gap between assigned and unassigned abilities. We compare two conditions on BBH: no statement about competence and a statement that the model is unreliable on unassigned abilities. We use four ability assignments and five questions per task type. Each question is run twice at temperature zero, and a given model receives the same questions in both conditions. Llama-3.2-1B is a weak-model reference and is tested only without the statement; the other five models are evaluated under both conditions.

The accuracy gap is the accuracy on assigned abilities minus the accuracy on unassigned abilities. The additional gap is the change in this quantity relative to the condition with no competence statement. If the prompt reliably enforced specialization, it would enlarge this gap by lowering accuracy on unassigned abilities while preserving accuracy on assigned abilities.

\begin{table}[ht]
\centering
\scriptsize
\caption{Results of ability assignment by system prompt on BBH.}
\label{tab:competence_full}
\resizebox{\linewidth}{!}{
\begin{tabular}{llrrrr}
\toprule
Model & Condition & \shortstack{Unassigned\\ability acc.} & \shortstack{Assigned\\ability acc.} & Accuracy gap & Additional gap \\
\midrule
DeepSeek-Chat & No statement & 0.812 & 0.800 & $-0.012$ & -- \\
& Told unreliable & 0.775 & 0.792 & 0.017 & 0.029 \\
\midrule
Llama-3.1-70B & No statement & 0.781 & 0.783 & 0.002 & -- \\
& Told unreliable & 0.750 & 0.767 & 0.017 & 0.015 \\
\midrule
GPT-4o-mini & No statement & 0.744 & 0.708 & $-0.035$ & -- \\
& Told unreliable & 0.675 & 0.667 & $-0.008$ & 0.027 \\
\midrule
GPT-OSS-120B & No statement & 0.869 & 0.825 & $-0.044$ & -- \\
& Told unreliable & 0.831 & 0.833 & 0.002 & 0.046 \\
\midrule
Qwen-2.5-7B & No statement & 0.494 & 0.558 & 0.065 & -- \\
& Told unreliable & 0.219 & 0.342 & 0.123 & 0.058 \\
\midrule
Llama-3.2-1B & No statement & 0.212 & 0.192 & $-0.021$ & -- \\
\bottomrule
\end{tabular}}
\end{table}

Declaring unassigned abilities unreliable increases the accuracy gap by $0.015$--$0.058$ across the five models tested under both conditions. Accuracy on assigned abilities also declines for four of the five models. These results suggest that the tested instructions affect overall response behavior without reliably restricting performance to the assigned abilities. The models can still solve tasks outside their assigned abilities.

In the main evaluation, each agent's declared abilities determine the model used for execution together with its healthy or degraded condition. A healthy agent uses the strong model for a declared ability and the weak model for a missing ability; a degraded agent uses the weak model. With complementary ability pairs, an incorrect route can therefore change the model applied to part of the task and affect task accuracy.

\section{Implementation Details}
\label{app:implementation}

\subsection{Full Procedure}
\label{app:algorithm}

\Cref{alg:meshheal} places routing, review, Takeover, detection, and reintegration in one procedure. The task state $X_t=(Q_t,T_t,R_t,C_t,Y_t,P_t)$ contains the query, task type, required and covered abilities, ordered accepted outputs, and routing history. Controller $i$ stores its inferred routing state $h_i\in\{\meshstate{normal},\meshstate{watched},\meshstate{isolated}\}$, recent evidence $W_i$ indexed by task type and ability, a cache $L_i$ of score and routing-state summaries received from peers, and Bellman--Ford distance $D_i(s)$ to each ability $s$.

In \textsc{Assess}, $H$ and $X$ denote the ability set and task state supplied by the caller, and $b_{\mathrm{probe}}$ is true for a recovery probe. The procedure returns the final score $q$ and the output $\widehat y$, which is the original output under \meshaction{Accept} and the Takeover synthesis under \meshaction{Takeover}. We use $\tau_{\mathrm{to}}=0.40$ for the low-score threshold and $\sigma_{\mathrm{com}}=0.55$ for the review-disagreement threshold.

The three collaboration actions specify how the task state changes. Handoff forwards the state unchanged; Contribute adds a reviewed partial result and expands $C_t$; and Finalize completes the remaining work and proposes a full answer. Review then chooses Accept or Takeover. For Finalize, the controller returns the accepted answer or its replacement. Review determines whether the current output is kept or replaced, while the inferred routing state affects later routes. The displayed prompts and traces use these same action names.

\begin{algorithm}[t]
\caption{\method{} task execution and routing adaptation}
\label{alg:meshheal}
\small
\begin{algorithmic}[1]
\Require Graph $G=(\mathcal V,\mathcal E)$, abilities $\{\mathcal K_i\}$,
task stream, parameters in \Cref{tab:parameters}

\State Initialize $h_i\gets\meshstate{normal}$,
$\mathcal K_i^{\mathrm{original}}\gets\mathcal K_i$,
$W_i,L_i\gets\emptyset$, $D_i(s)$,
$p_i^{\mathrm{probe}}\gets-\infty$

\For{each task $t$ with query $Q_t$ and type $T_t$}
    \State sample $i\sim\mathrm{Unif}(\mathcal V)$;
    $C_t,Y_t,P_t\gets\emptyset$;
    $b_{\mathrm{probe}}\gets\mathsf{false}$;
    $\widehat y_t\gets\bot$

    \If{$h_i=\meshstate{isolated}$}
        \State $i\gets\Call{RawRelay}{i,L_i,D_i}$
    \EndIf
    \State $R_t\gets\Call{InferAbilities}{i,Q_t,T_t}$

    \While{$\widehat y_t=\bot$ and budget remains}
        \State $M_t\gets R_t\setminus C_t$

        \If{$h_i=\meshstate{isolated}$ and not $b_{\mathrm{probe}}$}
            \State $i\gets\Call{RawRelay}{i,L_i,D_i}$;
            \textbf{continue}
        \EndIf

        \If{not $b_{\mathrm{probe}}$ and there exists probe-eligible
        $j\in\mathcal N_i$ with
        $\mathcal K_j^{\mathrm{original}}\cap M_t\neq\emptyset$}
            \State select one such $j$;
            $p_j^{\mathrm{probe}}\gets t$;
            $i\gets j$;
            $b_{\mathrm{probe}}\gets\mathsf{true}$;
            \textbf{continue}
        \EndIf

        \State $H_{i,t}\gets
        \mathcal K_i^{\mathrm{original}}\cap M_t$

        \If{$H_{i,t}=\emptyset$}
            \State $i\gets\Call{NextHop}{i,M_t,L_i,D_i}$;
            \textbf{continue}
        \ElsIf{$H_{i,t}\subsetneq M_t$}
            \State $a\gets\meshaction{Contribute}$
        \Else
            \State $a\gets\meshaction{Finalize}$
        \EndIf

        \State $X_t\gets(Q_t,T_t,R_t,C_t,Y_t,P_t)$
        \State $y\gets\Call{Generate}{i,a,X_t,H_{i,t}}$
        \State $(q_{i,t},\widehat y)
        \gets\Call{Assess}{i,y,H_{i,t},h_i,b_{\mathrm{probe}},X_t}$

        \State $h_i'\gets
        \Call{UpdateEvidence}{i,T_t,H_{i,t},q_{i,t},W_i,L_i}$

        \If{$h_i'\neq h_i$}
            \State $h_i\gets h_i'$
            \If{$h_i=\meshstate{isolated}$}
                \State $p_i^{\mathrm{probe}}\gets t$
            \EndIf
            \State advertise eligibility and distance changes
        \EndIf

        \State $b_{\mathrm{probe}}\gets\mathsf{false}$

        \If{$a=\meshaction{Contribute}$}
            \State append $\widehat y$ to $Y_t$;
            $C_t\gets C_t\cup H_{i,t}$;
            append $i$ to $P_t$
            \State $i\gets
            \Call{NextHop}{i,R_t\setminus C_t,L_i,D_i}$
        \Else
            \State $\widehat y_t\gets\widehat y$
        \EndIf
    \EndWhile

    \If{$\widehat y_t=\bot$}
        \State mark task $t$ unanswered
    \Else
        \State output $\widehat y_t$
    \EndIf
\EndFor
\end{algorithmic}
\end{algorithm}

\begin{algorithm}[t]
\caption{\method{} review and correction}
\label{alg:meshheal-recovery}
\fontsize{8.0}{8.35}\selectfont
\begin{algorithmic}[1]

\Procedure{Assess}{$i,y,H,h_i,b_{\mathrm{probe}},X$}
    \State obtain independent scores $u_1,u_2,u_3$
    from one eligible \meshstate{normal} reviewer
    with the required ability
    \State $q_0\gets\operatorname{median}(u_1,u_2,u_3)$;
    $s_u\gets\operatorname{sd}_{\mathrm{sample}}(u_1,u_2,u_3)$

    \If{$q_0\le\tau_{\mathrm{to}}$
        or $h_i=\meshstate{watched}$
        or $b_{\mathrm{probe}}$
        or $s_u\ge\sigma_{\mathrm{com}}$}

        \State select two distinct eligible reviewers A and B when possible;
        otherwise make two independent calls to the same eligible reviewer

        \State reviewers judge independently, exchange explanations,
        and judge again; let the final scores be $v_A,v_B$

        \State $q\gets(v_A+v_B)/2$

        \If{$q\le\tau_{\mathrm{to}}$}
            \State $\widehat y\gets
            \Call{TakeoverSynthesis}{A,B,y,X}$
            \State \Return $(q,\widehat y)$
        \EndIf
    \Else
        \State $q\gets q_0$
    \EndIf

    \State \Return $(q,y)$
\EndProcedure

\end{algorithmic}
\end{algorithm}

\textsc{UpdateEvidence} records $q_{i,t}$ for every $s\in H_{i,t}$,
updates the matched peer-relative shortfall $x_{i,t,s}$ when the required
summaries are available, and appends the resulting evidence to $W_i$.
It then applies \textsc{RelativeUpdate} to obtain the agent-level routing
state $h_i'$, following the detector rules in \Cref{sec:slow}.

\textsc{RawRelay} forwards the unchanged task state from an isolated agent, either the entry agent or an intermediate relay, toward a normal agent using $D_i$, without any LLM call.

\textsc{NextHop} reads only the cached eligibility states of the current controller's neighbors in $L_i$ and its distance table $D_i$. It samples a missing ability uniformly and routes toward its nearest holder using the cached Bellman--Ford distances. Normal and watched agents have identical execution eligibility; isolated agents are excluded as execution providers but may relay the task state without LLM calls.

\textsc{Assess} attributes $q_{i,t}$ to the agent that generated the original output under both Accept and Takeover. \textsc{RelativeUpdate} computes the shortfall $x_{i,t,s}$ only from observations with the same task type and ability as the current output; the normal range $m_{i,T}$ and $p^{25}_{i,T}$ pools the mean shortfalls of normal neighbors over all abilities within the same task type $T$. The resulting threshold decision updates one inferred routing state for the agent. The procedure also requires a minimum number of observations, checks for two consecutive crossings of the isolation threshold, and applies reversible routing-state transitions. Score summaries update neighbor caches, while eligibility and distance advertisements propagate from neighbor to neighbor. Probe timers are independent local events, so no coordinator scans the network. Later tasks use the latest local eligibility and distance tables.

\subsection{Agents and abilities}

\begin{table}[ht]
\centering
\small
\caption{BBH ability assignment in the six-agent setting.}
\label{tab:bbh_agent_skills}
\begin{tabular}{ll}
\toprule
Agent & Declared abilities \\
\midrule
A1 & reasoning, mathematical, knowledge \\
A2 & reasoning, mathematical, knowledge \\
A3 (degraded) & mathematical, knowledge, sequence, inference \\
A4 & language, sequence, spatial, inference \\
A5 (degraded) & reasoning, language, spatial \\
A6 & language, sequence, spatial, inference \\
\bottomrule
\end{tabular}
\end{table}

\begin{table}[ht]
\centering
\small
\caption{BBH task types and required abilities.}
\label{tab:task_skills}
\begin{tabular}{ll}
\toprule
Task type & Required abilities \\
\midrule
\texttt{causal\_judgement} & reasoning + inference \\
\texttt{formal\_fallacies} & reasoning + inference \\
\texttt{tracking\_shuffled\_objects\_five\_objects} & reasoning + sequence \\
\texttt{geometric\_shapes} & mathematical + spatial \\
\texttt{object\_counting} & mathematical + spatial \\
\texttt{date\_understanding} & mathematical + language \\
\texttt{ruin\_names} & language + knowledge \\
\bottomrule
\end{tabular}
\end{table}

\begin{table}[ht]
\centering
\scriptsize
\caption{MATH ability assignment.}
\label{tab:math_agent_skills}
\begin{tabularx}{\linewidth}{@{}lX@{}}
\toprule
Agent & Declared abilities \\
\midrule
A1 & \texttt{symbolic\_manipulation}, \texttt{numeric\_computation}, \texttt{theorem\_knowledge} \\
A2 & \texttt{symbolic\_manipulation}, \texttt{numeric\_computation}, \texttt{theorem\_knowledge} \\
A3 (degraded) & \texttt{numeric\_computation}, \texttt{theorem\_knowledge}, \texttt{enumeration}, \texttt{multi\_step\_deduction} \\
A4 & \texttt{enumeration}, \texttt{multi\_step\_deduction}, \texttt{problem\_translation}, \texttt{figure\_reasoning} \\
A5 (degraded) & \texttt{symbolic\_manipulation}, \texttt{problem\_translation}, \texttt{figure\_reasoning} \\
A6 & \texttt{enumeration}, \texttt{multi\_step\_deduction}, \texttt{problem\_translation}, \texttt{figure\_reasoning} \\
\bottomrule
\end{tabularx}
\end{table}

\begin{table}[ht]
\centering
\fontsize{6.4}{7.4}\selectfont
\caption{MATH subjects, required abilities, and assignment rationale.}
\label{tab:math_skills}
\begin{tabularx}{\linewidth}{@{}p{0.22\linewidth}p{0.45\linewidth}X@{}}
\toprule
Subject & Required abilities & Rationale \\
\midrule
\texttt{prealgebra} & \skillpair{numeric\_computation}{problem\_translation} & word-problem modeling + arithmetic \\
\texttt{algebra} & \skillpair{symbolic\_manipulation}{multi\_step\_deduction} & expression transformation + multi-step solving \\
\texttt{geometry} & \skillpair{numeric\_computation}{figure\_reasoning} & geometric relations + numerical calculation \\
\texttt{number\_theory} & \skillpair{symbolic\_manipulation}{enumeration} & algebraic constraints + systematic cases \\
\path{counting_and_probability} & \skillpair{symbolic\_manipulation}{enumeration} & counting expressions + complete case coverage \\
\texttt{intermediate\_algebra} & \skillpair{symbolic\_manipulation}{multi\_step\_deduction} & deeper transformations + longer deduction \\
\texttt{precalculus} & \skillpair{theorem\_knowledge}{figure\_reasoning} & function properties + graphical intuition \\
\bottomrule
\end{tabularx}
\end{table}

The MATH mapping separates operations on mathematical objects from the reasoning needed to formulate or complete a solution. Algebraic subjects pair symbolic transformation with multi-step deduction. Geometry and precalculus pair calculation or theorem knowledge with reasoning about figures, while subjects solved by considering separate cases pair symbolic constraints with enumeration. The assignments follow common stages of mathematical problem solving and associate each ability with a specific step in the solution process.

\begin{table}[ht]
\centering
\scriptsize
\caption{MMLU-Pro ability assignment.}
\label{tab:mmlu_agent_skills}
\begin{tabularx}{\linewidth}{@{}lX@{}}
\toprule
Agent & Declared abilities \\
\midrule
A1 & \texttt{humanities\_knowledge}, \texttt{science\_knowledge}, \texttt{quantitative\_methods} \\
A2 & \texttt{humanities\_knowledge}, \texttt{science\_knowledge}, \texttt{quantitative\_methods} \\
A3 (degraded) & \texttt{science\_knowledge}, \texttt{quantitative\_methods}, \texttt{scenario\_inference}, \texttt{stepwise\_analysis} \\
A4 & \texttt{scenario\_inference}, \texttt{stepwise\_analysis}, \texttt{text\_comprehension}, \texttt{structural\_reasoning} \\
A5 (degraded) & \texttt{humanities\_knowledge}, \texttt{text\_comprehension}, \texttt{structural\_reasoning} \\
A6 & \texttt{scenario\_inference}, \texttt{stepwise\_analysis}, \texttt{text\_comprehension}, \texttt{structural\_reasoning} \\
\bottomrule
\end{tabularx}
\end{table}

\begin{table}[ht]
\centering
\fontsize{6.4}{7.4}\selectfont
\caption{MMLU-Pro domains, required abilities, and assignment rationale.}
\label{tab:mmlu_skills}
\begin{tabularx}{\linewidth}{@{}p{0.18\linewidth}p{0.45\linewidth}X@{}}
\toprule
Domain & Required abilities & Rationale \\
\midrule
\texttt{health} & \skillpair{science\_knowledge}{text\_comprehension} & medical knowledge + comprehension of long questions \\
\texttt{history} & \skillpair{humanities\_knowledge}{scenario\_inference} & historical facts + contextual inference \\
\texttt{biology} & \skillpair{science\_knowledge}{structural\_reasoning} & biology knowledge + structural/process reasoning \\
\texttt{psychology} & \skillpair{humanities\_knowledge}{scenario\_inference} & theory + scenario judgment \\
\texttt{philosophy} & \skillpair{humanities\_knowledge}{scenario\_inference} & doctrine + argumentative stance \\
\texttt{economics} & \skillpair{quantitative\_methods}{text\_comprehension} & interpretation + model/calculation \\
\texttt{physics} & \skillpair{quantitative\_methods}{structural\_reasoning} & quantitative analysis + physical representation \\
\bottomrule
\end{tabularx}
\end{table}

The MMLU-Pro mapping pairs the main source of domain knowledge with the operation needed to apply it. Here, \texttt{structural\_reasoning} denotes reasoning about structures, processes, and physical representations. Knowledge-heavy domains require text comprehension, contextual inference, or reasoning about structure. Economics and physics combine quantitative methods with textual or structural interpretation. The two parts remain complementary: one agent supplies relevant knowledge or methods, and another applies them to the question.

The BBH assignments are inspired by the reasoning capabilities represented across BBH / BIG-Bench Extra Hard \citep{kazemi2025bbeh}; MATH and MMLU-Pro use names chosen for each dataset while preserving the same controlled structure. Within each dataset, exactly three agents hold each ability, and every task combines two abilities with disjoint sets of agents. Under this assignment, at least two agents are needed to cover both required abilities. Every ability held by degraded agents A3 and A5 is also held by two healthy agents after both targets degrade. The mapping defines complementary ability sets and preserves alternatives to each degraded agent.

At task entry, \textsc{Inferabilities} receives the complete seven-ability vocabulary, brief ability descriptions, the query and task type, and examples covering all evaluated task types and their required ability pairs. The entry LLM selects a pair and stores it in $R_t$; controllers then use this inferred pair for action selection and routing. Ability inference is therefore a constrained choice over the documented mappings in Tables~\ref{tab:task_skills}, \ref{tab:math_skills}, and~\ref{tab:mmlu_skills}.

\begin{table}[ht]
\centering
\small
\caption{Exact match between inferred ability pairs and the documented task-type mappings.}
\label{tab:skill_inference}
\begin{tabular}{lcccc}
\toprule
Dataset & BBH & MATH & MMLU-Pro & Overall \\
\midrule
Exact pair match & 100\% & 100\% & 100\% & 100\% \\
\bottomrule
\end{tabular}
\end{table}

\subsection{Models, Data, and Degradation}

\begin{table}[ht]
\centering
\small
\caption{Models used for task execution and induced degradation.}
\label{tab:models}
\begin{tabular}{ll}
\toprule
Execution condition & Model \\
\midrule
Declared ability, healthy agent & \texttt{openai/gpt-oss-120b} \\
Missing ability or degraded agent & \texttt{meta-llama/llama-3.2-1b-instruct} \\
\bottomrule
\end{tabular}
\end{table}

The main stream contains 504 tasks in 12 balanced blocks. Tasks 1--252 form the healthy phase.
Tasks 253--504 use different questions from the same task-type distribution and form the degraded
phase. At task 253, A3 and A5 switch from the strong model to the weak model while their declared
ability sets remain unchanged.

\begin{table}[ht]
\centering
\small
\caption{Datasets and scoring rules.}
\label{tab:datasets}
\begin{tabular}{ll}
\toprule
Dataset & Scoring rule \\
\midrule
BIG-Bench Hard & Exact answer extraction and matching \\
MATH & Mathematical equivalence with \texttt{math\_equal} \\
MMLU-Pro & Extracted option letter \\
\bottomrule
\end{tabular}
\end{table}

A missing or unparsable answer is counted as incorrect. Table~\ref{tab:skill_inference} shows that the LLM can accurately infer the abilities required for each task. Results for persistent degradation on each of BBH, MATH, and MMLU-Pro, together with the studies of alternative models and individual components, average five seeded task streams per setting. For the studies of degradation and recovery at different times, network scaling, and detection, the corresponding tables distinguish aggregated accuracies from agent counts; \Cref{tab:state_trajectory} reports a representative state trajectory.

The healthy and degraded phases preserve the task-type distribution and use disjoint questions, so the task mix remains comparable across distinct questions. Degradation is introduced by changing the model while preserving the agent identifier, declared ability set, and communication links. Task outputs and peer reviews provide the resulting evidence for routing-state updates.

\subsection{Baselines}

\begin{table}[ht]
\centering
\scriptsize
\caption{Baseline implementations.}
\label{tab:baselines}
\resizebox{\linewidth}{!}{
\begin{tabular}{p{0.16\linewidth}p{0.27\linewidth}p{0.48\linewidth}}
\toprule
Label & Source & Implementation used in our framework \\
\midrule
AgentNet & \citet{yang2025agentnet} & Native decentralized routing with adaptive edge weights and pruning. \\
RAPS & \citet{li2026raps} & Peer reputation updates local links and routing avoids low-reputation peers. \\
AutoGen & \citet{wu2024autogen} & A mandatory GroupChatManager-style hub delegates work, receives worker outputs, and produces the final answer. \\
SAC & \citet{lee2026sac} & Three teams produce answers, run two filtering and refinement rounds, and aggregate by majority. \\
Symphony & \citet{guan2026symphony} & GlobalLinUCB selects agents with the required abilities; three plans run in parallel; the final result is selected by voting and used to update routing statistics. \\
\bottomrule
\end{tabular}}
\end{table}

Each baseline uses the authors' released implementation. We retain its native routing, selection, filtering, and aggregation logic and connect it to a shared execution layer that fixes the task order, agent identities, ability assignments, models, degradation schedule, worker prompts, and Handoff--Contribute--Finalize task interface. The resulting comparison covers decentralized routing or reputation (AgentNet and RAPS), mandatory central delegation (AutoGen), and repeated candidate generation and aggregation (SAC and Symphony). We count every model call and all model tokens across routing, execution,
review, and aggregation, including repeated review samples and committee rounds.
Token cost includes both input prompt tokens and generated output tokens.

\subsection{Review, Accept, and Takeover}

For each Contribute or Finalize output, the controller samples one reviewer in the \meshstate{normal} state that holds the abilities used to produce it and calls that reviewer three times in independent contexts. \meshstate{Watched} and \meshstate{isolated} peers are ineligible throughout initial review and committee selection. Every call uses the same rubric for that type of output and the same score definitions shown in App.~\ref{app:prompts}. The median of the initial scores is $q_0$. Let $\bar u=(u_1+u_2+u_3)/3$. Disagreement among the three scores is measured by the sample standard deviation
$$
s_u=\sqrt{\frac{1}{3-1}\sum_{r=1}^{3}(u_r-\bar u)^2}.
$$
A two-reviewer committee is called when $q_0\leq0.40$, the agent is \meshstate{watched}, the output is a recovery probe, or $s_u\geq0.55$. All other outputs are accepted using the initial median score.

The controller selects distinct peers as Reviewer A and Reviewer B whenever two are available. If only one eligible reviewer is available, the controller uses that peer in two independent review contexts. The reviewers first score the output independently and explain their judgments. Each reviewer then sees both first-round scores and explanations and independently returns a second judgment. A score above $0.40$ leads to Accept. A score at most $0.40$ leads to Takeover: the two reviewers produce correction candidates, and a normal committee reviewer synthesizes them with the accepted task context. The controller replaces the original output with this synthesis. This replacement is not reviewed recursively; its cost is included, while the score attributed to the agent that generated the original output remains the detector evidence.

Mixture-of-Agents allows the same LLM to be reused within or across layers and to generate independently sampled candidates before aggregation \citep{wang2025moa}. We use this form of model reuse when only one eligible peer in the \meshstate{normal} state has the relevant ability. The controller calls that peer in two independent contexts labeled Reviewer A and Reviewer B. This preserves review and Takeover availability and provides two separately sampled judgments. Every reviewed output produces one detector score, attributed to the agent that generated the original output under both Accept and Takeover.

The two-timescale procedure connects correction at the fast timescale to detection at the slow timescale. One reviewer provides the initial check, and committee discussion is reserved for low-scoring or uncertain outputs, outputs from \meshstate{watched} agents, and recovery probes. Under Accept, the reviewed output remains in the task state. Under Takeover, the synthesized replacement takes its place. In both cases, the final review score remains attributed to the agent that generated the original output and supports later routing updates.

\textbf{Degradation detector.}
\label{app:health_detector}
Each output produced by agent $i$ on task $t$ receives a final review score $q_{i,t}$. For every ability $s\in H_{i,t}$ used in that output, the controller records $(T_t,s,q_{i,t})$ and maintains a separate detector for $(T_t,s)$. Let $\mathcal P_{i,T,s}(t)$ contain the direct neighbors of agent $i$ that are currently in the \meshstate{normal} state and have recent reviewed outputs for the same task type $T$ and ability $s$. Let $\mu_{j,T,s}(t)$ be neighbor $j$'s mean review score over these recent outputs. The reference score is
\[
\bar q_{-i,T,s}(t)
=
\frac{1}{|\mathcal P_{i,T,s}(t)|}
\sum_{j\in\mathcal P_{i,T,s}(t)}
\mu_{j,T,s}(t),
\]
and the shortfall for ability $s$ is
\[
x_{i,t,s}
=
\bar q_{-i,T_t,s}(t)-q_{i,t}.
\]

For each $(T,s)$, controller $i$ averages its latest eight matched shortfalls to obtain $\bar x_{i,T,s}$. For the second comparison, the controller uses the mean shortfalls maintained by its \meshstate{normal} neighbors for all abilities under the same task type $T$. Let
$$
\mathcal C_{i,T}(t)
=
\{(j,s'): j\in\mathcal N_i,\ h_j=\meshstate{normal},
\text{ and } \bar x_{j,T,s'} \text{ is available}\}.
$$
Thus, a neighbor may contribute multiple values, one for each ability $s'$ for which it has enough matched evidence. Let
$$
m_{i,T}
=
\operatorname{median}
\left\{
\bar x_{j,T,s'}:(j,s')\in\mathcal C_{i,T}(t)
\right\},
$$
and
$$
p^{25}_{i,T}
=
Q_{0.25}
\left(
\left\{
\bar x_{j,T,s'}:(j,s')\in\mathcal C_{i,T}(t)
\right\}
\right).
$$
The detector statistic is
$$
z_{i,T,s}
=
\frac{
\bar x_{i,T,s}-m_{i,T}
}{
\max\!\left(m_{i,T}-p^{25}_{i,T},\,0.10\right)
}.
$$

The first comparison is ability-specific, while the second comparison pools the mean shortfalls $\bar x_{j,T,s'}$ across abilities to estimate the normal range for task type $T$.
The current output updates the statistic for its own task type and ability, while the inferred routing state applies to the agent. After three matched observations, a \meshstate{normal} agent becomes \meshstate{watched} when $z_{i,T,s}\geq1.2$. A \meshstate{normal} or \meshstate{watched} agent becomes \meshstate{isolated} after $z_{i,T,s}\geq2.0$ on two consecutive updates with matched evidence; the counter resets when $z_{i,T,s}<2.0$. Recovery applies the thresholds in reverse: an \meshstate{isolated} agent moves to \meshstate{watched} when $z_{i,T,s}<2.0$, and a later \meshstate{watched} update with $z_{i,T,s}<1.2$ restores \meshstate{normal} status. The minimum of three matched observations applies separately to each agent, task type, and ability. If this minimum is not met, or either $\mathcal P_{i,T,s}(t)$ or $\mathcal C_{i,T}(t)$ is empty, the current routing state is retained.

Peer summaries use recent answers to different questions with the same task type and ability. The rolling window averages over question-level variation. After every matched review, the controller sends updated score and shortfall summaries to direct neighbors. Routing-state and distance summaries are sent when those values change so that neighboring controllers can update their routing tables.

Under the evaluated ability mappings, each produced output is associated with exactly one remaining required ability, i.e., $|H_{i,t}|=1$. Thus, each reviewed output updates one $(T,s)$ detector and one agent-level routing state.

\subsection{Parameters}

\begin{table}[ht]
\centering
\small
\caption{Main \method{} parameters.}
\label{tab:parameters}
\begin{tabular}{ll}
\toprule
Parameter & Value \\
\midrule
Committee reviewer slots & 2 \\
Committee deliberation rounds & 2 \\
{Takeover} correction candidates & 2 \\
Routing hop limit & 4 hops \\
Maximum agents per task & 6 \\
Maximum reviewed outputs per task & 3 \\
Evidence window per agent, task type, and ability & 8 observations \\
Isolation threshold & $z \ge 2.0$ (2 consecutive updates) \\
Watch threshold & $z \ge 1.2$ \\
Scale floor & 0.10 \\
Minimum observations & 3 \\
Minimum recovery-probe interval & 10 tasks \\
{Takeover} score threshold & $q \le 0.40$ \\
Initial review samples & 3 \\
Committee sample-s.d. trigger & $s_u\geq0.55$ \\
\bottomrule
\end{tabular}
\end{table}

The observation window and the requirement of two consecutive threshold crossings provide repeated evidence before isolation. The \meshstate{watched} state adds scrutiny before isolation, and the local probe timer controls how quickly fresh evidence can replace degraded observations after recovery. Each isolated agent's controller counts tasks since isolation or its last probe. After 10 tasks, the agent becomes probe-eligible; a neighbor schedules the probe when matching work next arrives, and the counter resets when the probe runs. Routing and review limits bound the amount of work per task. Low initial scores and high uncertainty trigger committee review and its additional cost; Takeover follows only when the committee's final score remains at or below $0.40$. Unless explicitly varied in an ablation, all rubrics and parameters for review and detection are fixed before evaluation and used unchanged across datasets, model pairs, and network sizes.

\subsection{Sparse Networks}

All reviewers, including committee members, are direct neighbors of the agent whose output they review. In the evaluated sparse graphs, every agent producing an ordinary task output has at least one neighbor in the \meshstate{normal} state with the same ability. Each agent selected for degradation has two such neighbors after both targets are excluded from reviewer selection, so its \meshstate{watched} outputs and recovery probes use distinct peers for the two reviewer contexts. Another agent can have only one eligible neighbor with the same ability after an agent with that ability degrades; in that case, the controller reuses that peer in the two independent reviewer contexts described above. Every agent is also within three hops of an agent in the \meshstate{normal} state for each required ability, leaving one hop of slack under the routing limit of four.

\begin{table}[ht]
\centering
\small
\caption{Sparse graph size. Edge counts treat each bidirectional connection as one undirected link.}
\label{tab:sparse_edges}
\begin{tabular}{rrrr}
\toprule
Agents & Sparse edges & Fraction of complete graph & Maximum ability distance \\
\midrule
96  & 228  & 5.00\% & 3 \\
288 & 705  & 1.71\% & 3 \\
576 & 1454 & 0.88\% & 3 \\
\bottomrule
\end{tabular}
\end{table}

The sparse graphs preserve the local conditions for review and routing: every agent has a reviewer with the same ability, and every required ability is available from a reachable healthy agent within the routing limit. The detector compares eligible neighbors in the \meshstate{normal} state. Edge density falls from $5.00\%$ at 96 agents to $0.88\%$ at 576 agents. The comparison between complete and sparse graphs changes connectivity while preserving these conditions.

\section{Detailed Results}
\label{app:results}

This appendix provides the detailed results summarized in \Cref{sec:experiments}. It first separates accuracy by dataset and reports inference cost. It then varies the model pair, degradation timing, and network topology before examining the committee and detector. Unless stated otherwise, all experiments use the task construction and cost accounting described in App.~\ref{app:implementation}.

\subsection{Main Results}

\begin{table}[ht]
\centering
\fontsize{7.5}{8.3}\selectfont
\setlength{\tabcolsep}{2.0pt}
\renewcommand{\arraystretch}{0.98}
\caption{Full persistent-degradation results, reported as mean $\pm$ standard deviation over five seeded task streams per dataset. Each stacked entry gives H above D for the healthy and degraded phases. The Mean column averages the three datasets within each seed, and Change is computed within seed before aggregation. Bold marks the best mean.}
\label{tab:full_main_results}
\begin{tabular*}{\linewidth}{@{\extracolsep{\fill}}lccccc@{}}
\toprule
Method & BBH & MATH & MMLU-Pro & Mean & Change \\
\midrule
\textbf{\method} & \shortstack{$.899\pm.027$\\$\mathbf{.900\pm.014}$} & \shortstack{$.821\pm.019$\\$\mathbf{.802\pm.021}$} & \shortstack{$.774\pm.016$\\$\mathbf{.814\pm.018}$} & \shortstack{$\mathbf{.831\pm.013}$\\$\mathbf{.839\pm.012}$} & $\mathbf{+.007\pm.012}$ \\
Symphony & \shortstack{$.896\pm.022$\\$.889\pm.007$} & \shortstack{$.818\pm.020$\\$.790\pm.022$} & \shortstack{$.766\pm.018$\\$.742\pm.026$} & \shortstack{$.827\pm.013$\\$.807\pm.014$} & $-.020\pm.018$ \\
SAC & \shortstack{$.896\pm.016$\\$.894\pm.007$} & \shortstack{$.794\pm.026$\\$.639\pm.045$} & \shortstack{$.778\pm.020$\\$.706\pm.034$} & \shortstack{$.822\pm.014$\\$.746\pm.021$} & $-.076\pm.026$ \\
RAPS & \shortstack{$.884\pm.029$\\$.740\pm.070$} & \shortstack{$.726\pm.034$\\$.627\pm.061$} & \shortstack{$.790\pm.024$\\$.591\pm.067$} & \shortstack{$.800\pm.019$\\$.653\pm.041$} & $-.147\pm.045$ \\
AgentNet & \shortstack{$.862\pm.035$\\$.668\pm.066$} & \shortstack{$.631\pm.041$\\$.476\pm.056$} & \shortstack{$.694\pm.034$\\$.528\pm.061$} & \shortstack{$.729\pm.023$\\$.557\pm.039$} & $-.172\pm.047$ \\
AutoGen & \shortstack{$.898\pm.013$\\$.525\pm.034$} & \shortstack{$.665\pm.030$\\$.214\pm.029$} & \shortstack{$.796\pm.019$\\$.532\pm.043$} & \shortstack{$.787\pm.014$\\$.424\pm.023$} & $-.363\pm.028$ \\
\bottomrule
\end{tabular*}
\end{table}

\method{} has the highest mean degraded-phase accuracy on all three datasets. Its aggregate accuracy is $0.831\pm0.013$ in the healthy phase and $0.839\pm0.012$ in the degraded phase, with a paired phase change of $+0.007\pm0.012$. On MATH, its mean changes from $0.821\pm0.019$ to $0.802\pm0.021$, while Symphony and SAC show mean changes of $-0.028$ and $-0.155$. BBH and MMLU-Pro have nonnegative mean changes. RAPS, AgentNet, and AutoGen show mean phase changes of $-0.147\pm0.045$, $-0.172\pm0.047$, and $-0.363\pm0.028$.

\begin{table}[ht]
\centering
\scriptsize
\setlength{\tabcolsep}{4pt}
\caption{Healthy- and degraded-phase cost per task, reported as mean $\pm$ standard deviation over five seeds after averaging datasets within seed. Tokens include both input and output tokens; their means and standard deviations are rounded to the nearest thousand.}
\label{tab:cost_full}
\begin{tabular}{lrrrr}
\toprule
Method & H calls & D calls & H tokens (k) & D tokens (k) \\
\midrule
\textbf{\method} & $12.4\pm0.2$ & $13.6\pm0.3$ & $51\pm3$ & $51\pm5$ \\
Symphony & $18.1\pm0.1$ & $23.4\pm0.1$ & $86\pm5$ & $115\pm9$ \\
SAC & $30.3\pm0.1$ & $39.9\pm0.6$ & $116\pm4$ & $178\pm13$ \\
RAPS & $8.2\pm0.5$ & $10.9\pm0.3$ & $39\pm5$ & $48\pm3$ \\
AgentNet & $7.8\pm0.7$ & $10.3\pm0.2$ & $34\pm4$ & $43\pm2$ \\
AutoGen & $6.1\pm0.0$ & $6.0\pm0.0$ & $15\pm0$ & $15\pm0$ \\
\bottomrule
\end{tabular}
\end{table}

After degradation onset, \method{}'s mean model calls increase from $12.4\pm0.2$ to $13.6\pm0.3$ per task because the system performs more review and rerouting, while mean total model tokens are $(51\pm3)$k and $(51\pm5)$k in the two phases. Symphony's mean token cost increases from $(86\pm5)$k to $(115\pm9)$k, and SAC's from $(116\pm4)$k to $(178\pm13)$k as they repeatedly generate and refine parallel candidates. RAPS and AgentNet have lower mean token costs together with mean accuracy changes of $-0.147\pm0.045$ and $-0.172\pm0.047$. AutoGen has both the lowest mean cost and the lowest mean degraded-phase accuracy when its mandatory manager degrades.

\subsection{Models and Recovery}
\label{sec:model_and_recovery}

\begin{table}[ht]
\centering
\scriptsize
\setlength{\tabcolsep}{3.5pt}
\caption{Model pair A: GPT-OSS-120B is the strong model and Gemma-3-4B is the weak model. Results are mean $\pm$ standard deviation over five seeded task streams; Change is paired within seed and tokens include both input and output tokens.}
\label{tab:model_pair_a}
\begin{tabular}{lrrrr}
\toprule
Method & Healthy acc. & Degraded acc. & Change & Tokens/task \\
\midrule
\textbf{\method} & $\mathbf{0.929\pm0.018}$ & $\mathbf{0.905\pm0.016}$ & $\mathbf{-0.024\pm0.014}$ & $56{,}687\pm2{,}431$ \\
Symphony & $0.917\pm0.020$ & $0.865\pm0.021$ & $-0.052\pm0.017$ & $100{,}685\pm4{,}782$ \\
SAC & $0.917\pm0.019$ & $0.829\pm0.030$ & $-0.088\pm0.026$ & $130{,}938\pm6{,}214$ \\
RAPS & $0.877\pm0.031$ & $0.623\pm0.064$ & $-0.254\pm0.055$ & $35{,}746\pm2{,}963$ \\
AgentNet & $0.857\pm0.037$ & $0.694\pm0.056$ & $-0.163\pm0.048$ & $36{,}978\pm3{,}284$ \\
AutoGen & $0.905\pm0.016$ & $0.734\pm0.041$ & $-0.171\pm0.038$ & $15{,}377\pm1{,}126$ \\
\bottomrule
\end{tabular}
\end{table}

\begin{table}[ht]
\centering
\scriptsize
\setlength{\tabcolsep}{3.5pt}
\caption{Model pair B: Llama-3.3-70B is the strong model and Llama-3.2-1B is the weak model. Results are mean $\pm$ standard deviation over five seeded task streams; Change is paired within seed and tokens include both input and output tokens.}
\label{tab:model_pair_b}
\begin{tabular}{lrrrr}
\toprule
Method & Healthy acc. & Degraded acc. & Change & Tokens/task \\
\midrule
\textbf{\method} & $0.790\pm0.025$ & $0.782\pm0.023$ & $\mathbf{-0.008\pm0.015}$ & $50{,}802\pm2{,}317$ \\
Symphony & $\mathbf{0.806\pm0.023}$ & $\mathbf{0.786\pm0.021}$ & $-0.020\pm0.017$ & $96{,}982\pm4{,}563$ \\
SAC & $\mathbf{0.806\pm0.022}$ & $0.754\pm0.032$ & $-0.052\pm0.027$ & $122{,}234\pm5{,}908$ \\
RAPS & $0.790\pm0.030$ & $0.571\pm0.061$ & $-0.219\pm0.053$ & $30{,}761\pm2{,}684$ \\
AgentNet & $0.774\pm0.038$ & $0.476\pm0.065$ & $-0.298\pm0.056$ & $29{,}487\pm2{,}792$ \\
AutoGen & $0.754\pm0.028$ & $0.349\pm0.042$ & $-0.405\pm0.038$ & $12{,}986\pm956$ \\
\bottomrule
\end{tabular}
\end{table}

The accuracy ranking changes with the model pair, while \method{} uses fewer tokens than Symphony under both pairs. Under pair A, \method{}'s mean degraded-phase accuracy is $0.905\pm0.016$, compared with $0.865\pm0.021$ for Symphony, and its mean token cost is 56,687 versus 100,685. Under pair B, the degraded-phase means are $0.782\pm0.023$ and $0.786\pm0.021$; \method{} has a smaller mean paired phase change ($-0.008\pm0.015$ versus $-0.020\pm0.017$) and uses $52\%$ of Symphony's mean tokens.

\Cref{tab:staggered} reports mean results over five seeds for the experiment with degradation and recovery at different times. All methods complete the full 630-task stream, with 126 tasks in each phase. The fully recovered mean accuracy is 0.857, equivalent to approximately 108 correct answers per 126-task phase. Across the 504 tasks after the first agent degrades, the reported mean pooled accuracy is 0.812 at 49k tokens per task, compared with 0.806 and 105k for Symphony. Mean accuracy rises from 0.770 when both targets are degraded to 0.841 after A5 recovers and 0.857 after both recover. RAPS and AgentNet reach 0.524 and 0.563 after both models are restored.

\begin{figure}[!t]
\centering
\small
\includegraphics[width=\linewidth,trim=7 8 7 8,clip]{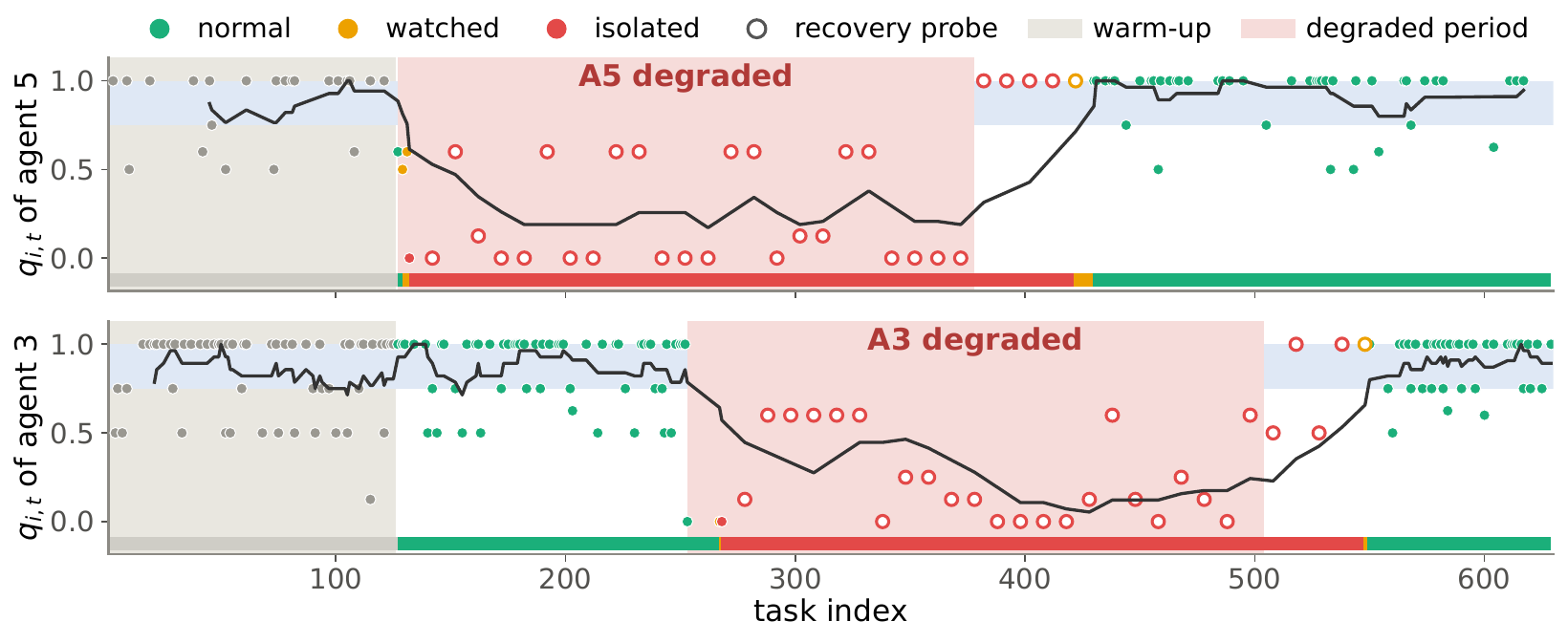}
\caption{Detection and reintegration under degradation and recovery at different times. Pink regions mark injected degradation. Each marker is one reviewed output, colored by the inferred state after the update; hollow markers are recovery probes. The bottom band shows the inferred state at each task, the curve shows the mean review score, and the blue region is the reference range for normal agents.}
\label{fig:detect_recover}
\end{figure}

\begin{table}[ht]
\centering
\scriptsize
\caption{Inferred routing-state changes in a representative run with degradation and recovery at different times.}
\label{tab:state_trajectory}
\begin{tabularx}{\linewidth}{l l X}
\toprule
Agent & True degraded interval & State sequence \\
\midrule
A5 & 127--378 & first evidence at task 3; \meshstate{normal} during warm-up $\rightarrow$ 129 \meshstate{watched} $\rightarrow$ 132 \meshstate{isolated} $\rightarrow$ 422 \meshstate{watched} $\rightarrow$ 430 \meshstate{normal} \\
A3 & 253--504 & first evidence at task 4; \meshstate{normal} during warm-up $\rightarrow$ 267 \meshstate{watched} $\rightarrow$ 268 \meshstate{isolated} $\rightarrow$ 548 \meshstate{watched} $\rightarrow$ 550 \meshstate{normal} \\

\bottomrule
\end{tabularx}
\end{table}

Warm-up is an evidence-collection phase in which all agents remain in ordinary routing. In the representative trajectory in \Cref{tab:state_trajectory}, A5 becomes watched at task 129 and isolated at task 132, five tasks after degradation begins. A3 becomes watched at task 267 and isolated at task 268, giving a detection delay of 15 tasks. Both remain isolated throughout their degraded intervals, during which they are observed only through recovery probes. After backend restoration, A5 and A3 move to watched at tasks 422 and 548 and return to normal routing at tasks 430 and 550, giving reintegration delays of 51 and 45 tasks, respectively.

\subsection{Larger Networks}

\Cref{tab:main_and_scale}(b) gives the accuracy and isolation results for complete and sparse graphs. Every degraded agent is \meshstate{isolated} in every topology. Complete-graph accuracy ranges from 0.808 to 0.827, and sparse-graph accuracy ranges from 0.803 to 0.818. The largest difference is 0.010 at 288 agents; the sparse graph is 0.006 higher at 576 agents. Expanding the task stream with the number of replicated six-agent ability profiles keeps detector evidence per agent comparable across scales. \Cref{tab:sparse_edges} gives the graph sizes and confirms that the maximum distance to a healthy agent with a required ability remains within the routing hop limit.

\subsection{Ablations}
\label{app:ablations}

\begin{table}[ht]
\centering
\small
\caption{Component study, reported as mean $\pm$ standard deviation over five seeded task streams. Phase change is paired within seed.}
\label{tab:component_ablation}
\begin{tabular}{lrrr}
\toprule
Metric & Full method & Single reviewer & No {Takeover} \\
\midrule
Healthy accuracy & $0.899\pm0.027$ & $0.909\pm0.022$ & $0.879\pm0.021$ \\
Degraded-phase accuracy & $\mathbf{0.900\pm0.014}$ & $0.865\pm0.029$ & $0.844\pm0.036$ \\
Paired phase change & $\mathbf{+0.001\pm0.016}$ & $-0.044\pm0.024$ & $-0.035\pm0.030$ \\
\bottomrule
\end{tabular}
\end{table}

The Single reviewer variant uses the initial reviewer with the same ability for the final judgment and any Takeover correction; it retains detection and rerouting while removing multi-reviewer discussion. The No Takeover variant retains review, committee escalation, detection, and rerouting but leaves a low-scoring output in the current task. The full configuration exceeds their mean degraded-phase accuracies by 0.035 and 0.056, respectively. No Takeover also lowers the healthy-phase mean by 0.020 because it leaves occasional low-quality outputs from healthy LLMs unchanged. The reported paired phase changes are $+0.001\pm0.016$, $-0.044\pm0.024$, and $-0.035\pm0.030$.

We next compare the detector based on peer comparisons with a fixed-threshold rule and a cumulative-sum (CUSUM) rule under the two model pairs \citep{page1954cusum}.

\begin{table}[ht]
\centering
\small
\caption{Detector performance across two BBH model pairs. Missed counts are out of two degraded agents and false-isolation counts are out of four healthy agents; each setting uses the full 504-task stream.}
\label{tab:detector_transfer}
\begin{tabular}{llrrrr}
\toprule
& & \multicolumn{2}{c}{Pair A} & \multicolumn{2}{c}{Pair B} \\
\cmidrule(lr){3-4}\cmidrule(lr){5-6}
Rule & Evidence & Missed & False & Missed & False \\
\midrule
Absolute threshold & Own scores & 0/2 & 4/4 & 0/2 & 0/4 \\
CUSUM & Own scores & 0/2 & 2/4 & 1/2 & 0/4 \\
Relative peer comparison & Scores relative to peers & \textbf{0/2} & \textbf{0/4} & \textbf{0/2} & \textbf{0/4} \\
\bottomrule
\end{tabular}
\end{table}

The absolute rule detects both degraded agents under both model pairs. Under pair A, it also isolates all four healthy agents. CUSUM reduces those false isolations to two and misses one degraded agent under pair B. Using the same review rubrics and detector parameters, relative peer comparison produces neither error under either model pair. Its reference adapts to peer scores for the same task type and ability while the decision thresholds remain fixed. The peer-relative reference relies on matched normal-neighbor summaries remaining representative of healthy performance.

The component study evaluates committee review and Takeover at the fast timescale. The detector comparison evaluates how peer-based evidence changes routing at the slow timescale under the two tested model pairs. Together, these studies support using review scores for both current-task correction and future routing.

\subsection{Review and {Takeover} Audit}
\label{app:review_audit}

We examine the initial review, committee reassessment, and Takeover across five seeded task streams. In the first two panels of \Cref{tab:agent_outcomes}, outputs from degraded agents are treated as positives and outputs from healthy agents as negatives. These labels describe the agent's injected condition, not the correctness of each output. A healthy agent can still produce an incorrect output, so the reported rates measure discrimination between injected conditions rather than detection of incorrect answers.

\begin{table}[htbp]
\centering

\begin{tabular}{@{}lc@{}}
\toprule
\multicolumn{2}{l}{\textbf{(a) Initial review}} \\
Metric & Value \\
\midrule
Mean score, healthy-agent outputs      & $0.89\pm0.01$ \\
Mean score, degraded-agent outputs     & $0.08\pm0.01$ \\
Recall on degraded-agent outputs       & $0.90\pm0.03$ \\
False-alarm rate on healthy-agent outputs & $0.053\pm0.007$ \\
Committee escalation rate for degraded-agent outputs & $0.98\pm0.02$ \\
\bottomrule
\end{tabular}

\vspace{1em}

\begin{tabular}{@{}lcc@{}}
\toprule
\multicolumn{3}{l}{\textbf{(b) Initial review and committee reassessment}} \\
Metric & Initial review & After committee \\
\midrule
Recall on degraded-agent outputs          & $0.90\pm0.03$ & $0.88\pm0.01$ \\
False-alarm rate on healthy-agent outputs & $0.053\pm0.007$ & $0.044\pm0.008$ \\
\midrule
Initial flags cleared, healthy-agent outputs  & -- & $0.33\pm0.07$ \\
Initial flags kept, degraded-agent outputs    & -- & $0.95\pm0.02$ \\
\bottomrule
\end{tabular}

\vspace{1em}

\begin{tabular}{@{}lc@{}}
\toprule
\multicolumn{2}{l}{\textbf{(c) Takeover outcomes}} \\
Outcome & Result \\
\midrule
\multicolumn{2}{l}{\emph{Finalize answers wrong before Takeover ($n=32$)}} \\
\quad corrected                       & $15/32$ ($47\%$) \\
\quad changed but still wrong         & $3/32$ ($9\%$) \\
\quad left unchanged                  & $14/32$ ($44\%$) \\
\quad corrected, among those changed  & $15/18$ ($83\%$) \\
\addlinespace
\multicolumn{2}{l}{\emph{Finalize answers correct before Takeover ($n=32$)}} \\
\quad changed to incorrect            & $1/32$ ($3\%$) \\
\addlinespace
\multicolumn{2}{l}{\emph{Takeover of Contribute outputs ($n=137$)}} \\
\quad task correct at the end         & $86\%$ \\
\bottomrule
\end{tabular}

\caption{Agent review performance and Takeover outcomes.}
\label{tab:agent_outcomes}
\end{table}

The initial review assigns mean scores of $0.89\pm0.01$ and $0.08\pm0.01$ to outputs from healthy and degraded agents, respectively. Each seed contains 31--33 outputs from degraded agents in this audit, of which 2--4 are missed by the initial low-score decision. The committee escalation rate is higher because \meshstate{watched} status or disagreement among the initial scores can also trigger committee review.

After committee reassessment, recall changes from $0.90\pm0.03$ to $0.88\pm0.01$, while the false-alarm rate decreases from $0.053\pm0.007$ to $0.044\pm0.008$. Among initially flagged outputs, the committee clears $0.33\pm0.07$ of healthy-agent cases and retains $0.95\pm0.02$ of degraded-agent cases.

Finalize answers can be checked directly against the gold answer. Among 32 initially wrong answers entering Takeover, 15 are corrected. The synthesis changes 18 of these answers, and 15 of the 18 changes are correct; the other 14 wrong answers are left unchanged. Among 32 initially correct answers entering Takeover, only one is changed to an incorrect answer.

A Contribute output cannot be scored directly from the final gold answer. Among the 137 tasks where Takeover replaces a Contribute output, $86\%$ end with a correct final answer. This number reports the final task outcome and is not a direct measure of intermediate
output correctness.

\section{Runtime Prompts}
\label{app:prompts}

This appendix presents representative runtime prompts for communicating ability matches to the LLM and reviewing outputs. Shaded cards distinguish prompt instructions from explanatory prose. \Cref{alg:meshheal} specifies the surrounding controller logic. For presentation, we standardize ability terminology and the action names Handoff, Contribute, and Finalize. Executable JSON field names, other wording, thresholds, and placeholders are retained; line breaks are adjusted to fit the page.

\subsection{Review Prompts}
\label{app:review_prompt}

Review depends on the task action. A Contribute reviewer evaluates the assigned intermediate output. A Finalize reviewer evaluates the finalizer's own output and whether it commits a concrete answer. The prompt identifiers use the same Contribute and Finalize names as the method. In both cases, the controller selects a reviewer that holds the relevant ability. Process issues such as exceeding the assigned scope or ignoring a draft remain advisory unless they create a concrete correctness error. Accept and Takeover are controller actions derived from the resulting score.

The cards retain the executed instructions with ability terminology and action labels standardized for presentation. The assigned part refers to work requiring one of the task's two abilities; \texttt{draft-half} refers to the earlier output covering the other ability. Executable JSON field names are unchanged.

\begin{promptcard}{Contribute judge system prompt (\texttt{COMMITTEE\_CONTRIBUTE\_JUDGE\_SYSTEM})}
You review a DRAFTER's INTERMEDIATE contribution in a multi-agent system. The
drafter was delegated just ONE part of a task requiring two abilities; it
should Contribute that half and leave the other half for Handoff to a teammate -
it should NOT Finalize (that needs the other ability too). You give one noisy
review signal; do not solve, repair, or decide degradation. Judge ONLY the
drafter's half, using only a ability you actually hold (see
REVIEWER_COMPETENCE). Return only valid JSON.
\end{promptcard}

The scoring schema anchors the returned signal and requires every low score to identify a concrete correctness error. The controller assigns reviewers who hold the relevant ability. If a reviewer does not hold that ability, the prompt directs it to return a neutral score with low confidence.

\begin{promptcard}{Contribute judge scoring schema}
Score each in [0,1]:
- half_correct: ONLY IF you are competent at the drafter's ability - re-
  derive that half; is it correct? LOW for a concrete error (miscalculation,
  invalid step, contradiction with the given facts/objects). If you are NOT
  competent at that ability, set 0.5 and lower confidence.
- in_scope: did the drafter STAY IN ITS LANE - do its half and leave the final decision OPEN?

SCORING RULE - normal_working_score MUST equal half_correct: score the
CORRECTNESS of the delegated half only. in_scope is ADVISORY: it must NOT lower
normal_working_score.
anchors: 1.00 half is sound / 0.50 cannot verify or weak / 0.25 concrete error /
0.00 unusable garbage for the delegated half.

Return ONLY valid JSON:
{ "half_correct": .., "in_scope": .., "normal_working_score": .., "confidence": ..,
  "specific_failure_found": bool, "failure_type": "none|wrong_half|over_reach|other",
  "failure_evidence": "quote from the drafter output; empty if none",
  "one_sentence_reason": ".." }
Constraints: normal_working_score<=0.40 => specific_failure_found=true + non-
empty evidence of a CORRECTNESS failure; over_reach alone NEVER sets
specific_failure_found.
\end{promptcard}

\begin{promptcard}{Finalize judge system prompt (\texttt{COMMITTEE\_FINALIZE\_JUDGE\_SYSTEM})}
You review a FINALIZER's answer in a multi-agent system. The finalizer did its
OWN part and must COMBINE it with the draft-half already in the trajectory to
commit ONE concrete final answer. The draft-half was produced by a teammate
using a ability the finalizer (and you) do NOT hold: treat it as GIVEN - this
review does not score whether the draft-half is right. Judge only the
finalizer's own half and how faithfully it combined. You give one noisy review
signal; do not solve/repair. Return only valid JSON.
\end{promptcard}

Finalize returns one concrete, scoreable answer; if no concrete answer is given, the score is at most 0.25. It uses the same numerical anchors and the same requirement to identify a concrete error as Contribute, applied to the finalizer's work within its declared ability and the committed answer. Draft handling affects the score when it creates a concrete error in the finalizer's own work or in the combined answer.

\subsection{ability Prompts}

Before execution, the entry LLM infers the required ability pair from the complete dataset ability vocabulary, brief descriptions, and examples mapping all evaluated task types to ability pairs. The controller stores this pair in the task state. It then compares the inferred pair with the agent's declared abilities and inserts one of three directives into the execution prompt. The evaluation selects the strong or weak model from the ability match and the agent's injected condition; the directive communicates the ability match to the LLM.

\begin{promptcard}{Complete competence match}
[COMPETENCE VERDICT - DECIDED BY SYSTEM, BINDING]
This task requires: {req}. You ARE competent at all of them. Solve it correctly and carefully.
On the FIRST line output exactly 'COMPETENCE_MATCH: yes', then give your answer.
\end{promptcard}

\begin{promptcard}{Partial competence match}
[COMPETENCE NOTE - PARTIAL FIT]
You can reliably handle only the parts of this task that fall WITHIN YOUR OWN abilities...
- If you select Finalize: you MUST commit a single concrete final answer,
  combining the part you can do with whatever the progress already provides for the rest.
- If you are only contributing an INTERMEDIATE part (a Contribute draft): do ONLY the part your
  abilities cover, state that partial result, and request Handoff for the remaining work.
On the FIRST line output exactly 'COMPETENCE_MATCH: partial', then give your answer/contribution.
\end{promptcard}

\begin{promptcard}{No competence match}
[COMPETENCE NOTE]
This task requires: {req}, none of which are your strengths. Answer as best you can,
but treat your own answer here as unreliable.
On the FIRST line output exactly 'COMPETENCE_MATCH: no', then give your answer.
\end{promptcard}

The no-match directive applies to generation assigned outside an agent's declared abilities. In \method{}, an agent with none of the remaining required abilities instead performs controller-only Handoff. App.~\ref{app:competence} reports the limited increase in the accuracy gap between assigned and unassigned abilities under the tested prompts.

\clearpage
\section{Case Studies}
\label{app:cases}

We present four representative execution traces showing collaboration across complementary abilities, multi-hop routing around an \meshstate{isolated} agent, committee review and Takeover after an incorrect output, and correction of a localized reasoning error. For presentation, we standardize ability terminology and action labels to match Sections~\ref{sec:method} and~\ref{app:implementation}. Recorded answers, numerical values, and executable field names are retained; logging metadata not needed to follow the execution is omitted.

\subsection{Collaboration and Routing}

\begin{casecard}{Case 1: Complementary abilities across two agents}
\textbf{Task:} \texttt{object\_counting}.\\
\textbf{Required abilities:} mathematical + spatial.\\
\textbf{Route:} A4 $\rightarrow$ A1.\\
\textbf{Outcome:} success; gold answer 15.\\[2pt]
\textbf{Question:} \emph{``I have a drum, four pianos, two accordions, a carrot,
a stalk of celery, three flutes, a cauliflower, a violin, a clarinet, a trombone,
and two trumpets. How many musical instruments do I have?''}
\end{casecard}

A4 is assigned the spatial ability but not the mathematical ability. It contributes the part corresponding to its assigned ability and leaves the remaining work to another agent.

\begin{tracecard}{A4 Contribute: Filter and enumerate}
RESULT: Based on the provided list, the musical instruments and their quantities are:
- Drum: 1
- Piano: 4
- Accordion: 2
- Flute: 3
- Violin: 1
- Clarinet: 1
- Trombone: 1
- Trumpet: 2

Non-instrument items (carrot, celery, cauliflower) are excluded.
The arithmetic summation can be completed by a teammate.
\end{tracecard}

The spatial component is now covered, while the mathematical component remains. The controller therefore routes the updated task state to A1, which holds the missing ability and is in the normal routing state.

\begin{tracecard}{Controller routing: A4 to A1}
Covered ability: spatial
Remaining ability: mathematical
NEXT_AGENT_ID: 1
\end{tracecard}

A1 receives the accepted output and completes the remaining operation.

\begin{tracecard}{A1 Finalize: Complete the mathematical component}
DESCRIPTION: Using the counts from Agent 4:
1 + 4 + 2 + 3 + 1 + 1 + 1 + 2 = 15.

RESULT: 15
\end{tracecard}

\begin{takeawaycard}{What this case shows}
The task is solved through complementary outputs from two agents. A4 performs the spatial
filtering and enumeration, routing carries that result, and A1 uses the remaining mathematical
ability in the Finalize step. The trace illustrates how
declared abilities determine both the division of work and the next routing decision.
\end{takeawaycard}

\begin{casecard}{Case 2: Multi-hop routing around an \meshstate{isolated} agent}
\textbf{Task:} \texttt{number\_theory}.\\
\textbf{Required abilities:} symbolic manipulation + enumeration.\\
\textbf{Network:} 288 agents, 705 bidirectional links, average degree 4.9, with
$1.7\%$ as many edges as the corresponding complete graph.\\
\textbf{Outcome:} success; gold answer 1575.\\[2pt]
\textbf{Question excerpt:} \emph{``A school has between 150 and 200 students...
What is the sum of all possible numbers of students enrolled?''}
\end{casecard}

Agent 61 first derives the candidate values needed for the remainder of the solution.

\begin{tracecard}{Agent 61 Contribute: Symbolic manipulation}
Assigned ability: symbolic manipulation

RESULT: 151, 157, 163, 169, 175, 181, 187, 193, 199
\end{tracecard}

The symbolic-manipulation component is now covered, while enumeration remains. Agent 12, the nearest holder of enumeration, is isolated, so its ability set is empty for routing. The updated Bellman--Ford distances lead to Agent 82, the nearest \meshstate{normal} holder of enumeration, three hops away.

\begin{tracecard}{Controller-only relays in a sparse network}
Remaining ability: enumeration
Agent 61 -> Agent 49    (routing after Contribute)
Agent 49 -> Agent 151   (Handoff)
Agent 151 -> Agent 82   (Handoff)
\end{tracecard}

The intermediate agents relay the task state without asking an LLM to generate task content.

\begin{tracecard}{Agent 82 Finalize: Complete the enumeration component}
Assigned ability: enumeration

DESCRIPTION: Add the nine values provided by Agent 61:
151 + 157 + 163 + 169 + 175 + 181 + 187 + 193 + 199 = 1575.

RESULT: 1575
\end{tracecard}

\begin{takeawaycard}{What this case shows}
Agent 61 contributes the symbolic-manipulation component. Local routing-state information then prevents the task from being assigned to an \meshstate{isolated} agent. Bellman--Ford distance updates guide two controller-only relays to Agent 82, which holds the remaining enumeration ability and uses Finalize to complete the task. The route uses only local eligibility and distance information.
\end{takeawaycard}

\subsection{Review and Takeover}

\begin{casecard}{Case 3: Fluent but incorrect output, committee reassessment, and Takeover}
\textbf{Task:} \texttt{geometric\_shapes}.\\
\textbf{Required abilities:} mathematical + spatial.\\
\textbf{Degraded agent:} A5.\\
\textbf{Outcome:} success; gold answer (J), triangle.\\[2pt]
\textbf{Question:} SVG path
\texttt{'M 30.17,45.97 L 58.79,40.36 L 18.10,15.70 M 18.10,15.70 L 30.17,45.97'}.
\end{casecard}

A5 remains responsive after degradation and produces a well-formed output. The failure is
therefore not visible from availability or surface form alone.

\begin{tracecard}{Step 1: A5 produces a fluent but incorrect contribution}
RESULT: 6 vertices. The segments connect in the order A-B-C-D-E-F-A,
forming a closed contour. This describes a hexagon.
\end{tracecard}

The path contains only three distinct vertices, so the output contains a concrete correctness error. The output is escalated to committee review. Committee members first assess the same output independently.

\begin{tracecard}{Step 2: Independent committee judgments}
Reviewer A: score 0.0

Reason: "The path contains only three distinct points and three line segments,
not six vertices forming a hexagon."

Reviewer B: score 0.5

Reason: "The shape classification is incorrect, but the vertex interpretation
requires reassessment."
\end{tracecard}

The first-round judgments disagree. Each reviewer then receives the other reviewer's assessment
and explanation before independently reassessing the original output.

\begin{tracecard}{Step 3: Committee reassessment}
First-round scores:  [0.0, 0.5]
Second-round scores: [0.0, 0.0]
Final review score:  0.0
Decision: Takeover
\end{tracecard}

After reassessment, both reviewers identify the output as incorrect. The final low score
remains attributed to A5 as evidence for routing-state updates, while Takeover repairs the current task immediately.

\begin{tracecard}{Step 4: Takeover correction}
The path contains three distinct vertices:
(30.17, 45.97), (58.79, 40.36), and (18.10, 15.70).

They are connected by three straight-line segments forming a closed
three-sided polygon: a triangle. No additional vertices or segments are present.
\end{tracecard}

The corrected output is placed in the task state, and the downstream Finalize step returns
(J), matching the gold answer.

\begin{takeawaycard}{What this case shows}
Degradation need not produce malformed or obviously unusable text: A5 returns a fluent and
confident output containing a substantive error. Independent reviewers initially disagree,
but committee reassessment resolves the disagreement. Takeover then replaces the incorrect
output before it propagates to the final answer, while the review score remains available as
evidence for subsequent routing-state updates.
\end{takeawaycard}

\begin{casecard}{Case 4: A localized error inside an otherwise coherent solution}
\textbf{Task:} \texttt{tracking\_shuffled\_objects\_five\_objects}.\\
\textbf{Required abilities:} reasoning + sequence.\\
\textbf{Degraded agent:} A5.\\
\textbf{Outcome:} success; gold answer (C), brown.
\end{casecard}

In this example, the overall response is coherent, but one state transition is incorrect. A5 state is watched. Its output therefore receives mandatory committee review.

\begin{tracecard}{A5 Contribute: Reason through the state updates}

1. Claire and Alice swap:
   Claire (red), Alice (brown), Bob (green), Dave (black), Eve (orange)

2. Dave and Eve swap:
   Claire (red), Alice (brown), Bob (green), Dave (orange), Eve (black)

3. Claire and Eve swap:
   Claire (black), Alice (brown), Bob (green), Dave (orange), Eve (red)

4. Bob and Claire swap:
   Claire (green), Alice (brown), Bob (black), Dave (orange), Eve (red)

5. Alice and Claire swap:
   Claire (green), Alice (black), Bob (brown), Dave (orange), Eve (red)

Therefore, Claire has the green ball.
\end{tracecard}

The first four state transitions are correct. The fifth is not: immediately before the final swap,
Claire holds green and Alice holds brown, so exchanging their balls should leave Claire with brown
and Alice with green.

Both committee reviewers then identify the incorrect transition.

\begin{tracecard}{Committee review identifies the failing step}
Reviewer A: The final swap states that Claire remains green, but replaying the
exchange gives Claire brown.

Reviewer B: Step 5 is the first incorrect state update; the preceding four
transitions are consistent.
\end{tracecard}

The committee's final score is at most $\tau_{\mathrm{to}}$, so Takeover replaces the output.

\begin{tracecard}{Takeover correction}
After the final Alice--Claire swap:
Claire holds brown and Alice holds green.
\end{tracecard}

The corrected output is placed in the task state, and the downstream Finalize step by an agent holding the sequence ability returns (C), matching the gold answer.

\begin{takeawaycard}{What this case shows}
The failure is subtle: four consecutive state updates are correct and only the final transition is
wrong. The response otherwise has the structure and fluency of a valid solution. The review
identifies the specific correctness error, and Takeover repairs the current task without
restarting the full task trajectory. This complements Case 3, where committee
reassessment resolves disagreement over a more visible semantic error.
\end{takeawaycard}

\end{document}